\documentclass[10pt,journal,compsoc]{IEEEtran}
\ifCLASSOPTIONcompsoc
  \usepackage[nocompress]{cite}
\else
  \usepackage{cite}
\fi
\ifCLASSINFOpdf
\else
\fi
\usepackage{amsmath,amssymb,amsfonts}
\usepackage{subcaption}
\usepackage{booktabs}
\usepackage{multirow}
\usepackage{colortbl} 
\usepackage{arydshln}
\usepackage{graphicx}
\usepackage{amsmath}
\usepackage{amssymb}
\usepackage{amsmath,amssymb,amsfonts}
\usepackage{algorithm}
\usepackage{booktabs}
\usepackage{algorithmic}
\usepackage{bm}
\usepackage{multirow}
\usepackage{amssymb}
\usepackage{subcaption}
\usepackage{mathtools}
\usepackage{makecell}
\usepackage{enumitem}
\usepackage{caption}
\usepackage{graphicx}
\usepackage{amsmath}
\usepackage{amsthm}
\usepackage{booktabs}
\usepackage{algorithm}
\usepackage{algorithmic}
\usepackage{graphicx}
\usepackage{amsmath}
\usepackage{amsmath,amssymb,amsfonts}
\usepackage{algorithm}
\usepackage{booktabs}
\usepackage{algorithmic}
\usepackage{bm}
\usepackage{multirow}
\usepackage{amssymb}
\usepackage{color}
\usepackage{subcaption}
\usepackage{multirow}
\usepackage{amssymb}
\usepackage{subcaption}
\usepackage{mathtools}
\usepackage{makecell}
\usepackage{enumitem}
\usepackage{caption}
\usepackage{algorithm} 
\begin{document}
%
\title{Online Knowledge Distillation via Mutual Contrastive Learning for Visual Recognition}
%
%
%
%

\author{Chuanguang~Yang,
        Zhulin~An,
        Helong~Zhou,
        Fuzhen~Zhuang,
        Yongjun~Xu,
        and Qian~Zhang
\IEEEcompsocitemizethanks{\IEEEcompsocthanksitem Chuanguang Yang is with Institute of Computing Technology, Chinese Academy of Sciences, Beijing 100190, China and University of Chinese Academy of Sciences, Beijing 100049, China. Email: yangchuanguang@ict.ac.cn.
\IEEEcompsocthanksitem Zhulin An and Yongjun Xu are with Institute of Computing Technology, Chinese Academy of Sciences, Beijing 100190, China. Email: \{anzhulin, xyj\}@ict.ac.cn.
\IEEEcompsocthanksitem Fuzhen Zhuang is with Institute of Artificial Intelligence, Beihang University, Beijing, China and Zhongguancun Laboratory, Beijing, China. Email: zhuangfuzhen@buaa.edu.cn.
\IEEEcompsocthanksitem Helong Zhou and Qian Zhang are with Horizon Robotics, Beijing, China. Email: \{helong.zhou, qian01.zhang\}@horizon.ai.
}
\thanks{(Corresponding author: Zhulin An.)}}

%
%

\markboth{Journal of \LaTeX\ Class Files,~Vol.~14, No.~8, August~2015}%
{Shell \MakeLowercase{\textit{et al.}}: Bare Demo of IEEEtran.cls for Computer Society Journals}
\IEEEtitleabstractindextext{%
\begin{abstract}
	  The teacher-free online Knowledge Distillation (KD) aims to train an ensemble of multiple student models collaboratively and distill knowledge from each other. Although existing online KD methods achieve desirable performance, they often focus on  class probabilities as the core knowledge type, ignoring the valuable feature representational information. We present a Mutual Contrastive Learning (MCL) framework for online KD. The core idea of MCL is to perform mutual interaction and transfer of contrastive distributions among a cohort of networks in an online manner. Our MCL can aggregate cross-network embedding information and maximize the lower bound to the mutual information between two networks. This enables each network to learn extra contrastive knowledge from others, leading to better feature representations, thus improving the performance of visual recognition tasks. Beyond the final layer, we extend MCL to intermediate layers and perform an adaptive layer-matching mechanism trained by meta-optimization.  Experiments on image classification and transfer learning to visual recognition tasks show that layer-wise MCL can lead to consistent performance gains against state-of-the-art online KD approaches. The superiority demonstrates that layer-wise MCL can guide the network to generate better feature representations. Our code is publicly avaliable at https://github.com/winycg/L-MCL.
\end{abstract}

}

\maketitle

\IEEEdisplaynontitleabstractindextext

%
\IEEEpeerreviewmaketitle

\IEEEraisesectionheading{\section{Introduction}\label{sec:introduction}}

%
%
%
%

\IEEEPARstart{D}{eep} Convolutional Neural Networks (CNNs) have achieved desirable performance over the past decade across a broad range of computer vision tasks, including image classification~\cite{he2016deep,huang2017densely,yang2020gated}, object detection~\cite{ren2016faster,he2017mask} and semantic segmentation~\cite{yang2022cross}. The high-performance models often require a large amount of computational and storage resources. The drawback limits these superior models to be deployed over resource-limited edge devices. To overcome the problem, many works aim to construct smaller yet more accurate CNNs. Mainstream model compression solutions are divided into \emph{pruning}~\cite{yang2019multi,cai2022prior}, \emph{parameter quantization}~\cite{gholami2021survey} and \emph{knowledge distillation}~\cite{hinton2015distilling,yang2022mixskd}. This paper focuses on Knowledge Distillation (KD) to improve network performance.

KD provides an intuitively effective paradigm to improve a small student network by absorbing knowledge from a large teacher network with better performance. Hinton'KD~\cite{hinton2015distilling} formulates the class probability distributions from the teacher network as soft labels to supervise the student's predictions. Based on this idea, some KD methods attempt to guide the student to mimic the teacher's meaningful knowledge, such as feature maps~\cite{romero2014fitnets} and refined information~\cite{zagoruyko2016paying,yang2021hierarchical,yang2022knowledge}.  The traditional KD follows a two-stage training pipeline and needs to pre-train a large teacher network. However, we may not have a sizeable pre-trained model readily at hand.

To overcome this issue, online KD~\cite{zhang2018deep,zhu2018knowledge,song2018collaborative} is proposed to prompt teacher-free distillation by training two or more student networks simultaneously. The idea of online KD is to take advantage of collaborative learning among multiple student networks. \emph{Deep mutual learning} (DML)~\cite{zhang2018deep} demonstrates that a cohort of models can benefit from mutual learning of \emph{class probability distributions}, \emph{i.e.} the final predictions. Each model in such a peer-teaching manner learns better than learning alone in conventional supervised training. Existing online KD works~\cite{zhang2018deep,zhu2018knowledge,song2018collaborative,chen2020online,guo2020online,wu2021peer} often focus on outcome-driven distillation by minimizing the final predictions among peer networks with various strategies. These methods ignore distilling feature information that is also valuable for online KD.

Unlike the class posterior, feature embeddings contain structured knowledge and are more tractable to capture dependencies among various networks. Although the previous AFD~\cite{chung2020feature} attempts to align intermediate feature maps in an online manner, Zhang \emph{et al.}~\cite{zhang2018deep} points out this would diminish the cohort diversity and harm the ability of mutual learning. To deal with feature embeddings meaningfully, we think a more desirable approach is contrastive learning from the perspective of visual representation learning.

Contrastive learning has been widely demonstrated as an effective framework for learning feature representations~\cite{schroff2015facenet,khosla2020supervised}, especially in recent self-supervised learning~\cite{he2020momentum,chen2020simple}. The core idea of contrastive learning is to pull positive pairs together and push negative pairs apart in the feature embedding space via a contrastive loss. The success behind contrastive learning is learned good features that are preferable to downstream visual recognition tasks~\cite{he2020momentum,chen2020simple,misra2020self}. From the perspective of mutual learning on class probabilities, we hypothesize that it may be desirable to perform \emph{Mutual Contrastive Learning} (MCL) among a cohort of models. Benefiting from MCL, it makes sense to take advantage of collaborative learning for better visual representation learning of each network, thus improving the recognition performance. 
\begin{figure}[t]
	\centering 
	
	\begin{subfigure}[t]{0.4\textwidth}
		\centering
		\includegraphics[width=\textwidth]{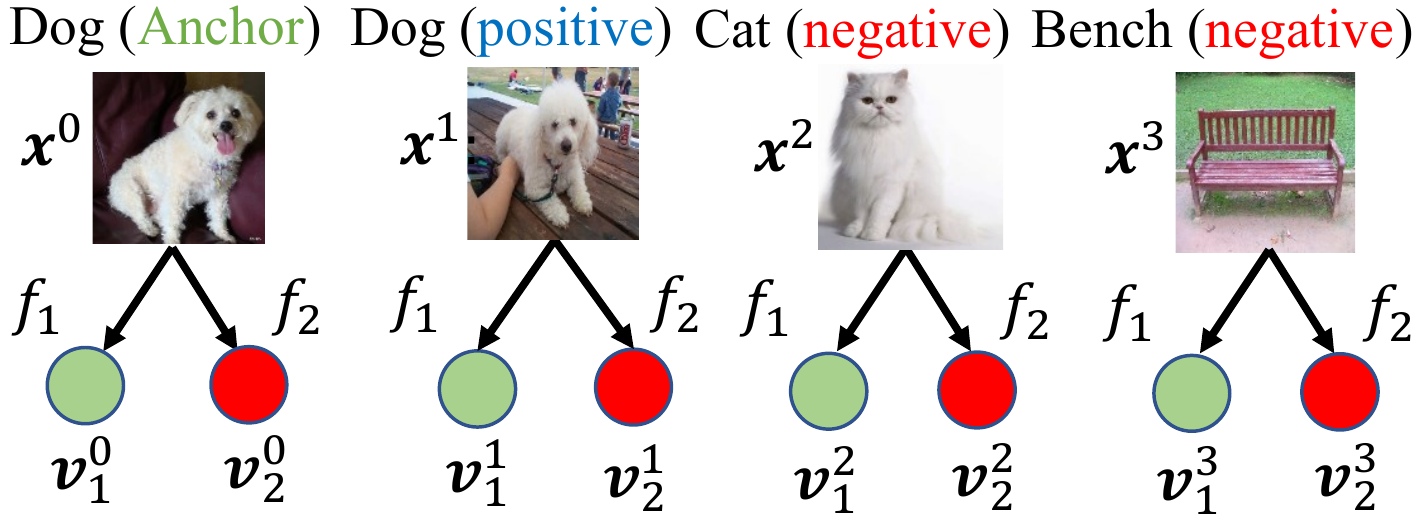}
		\caption{Positive and negative pairs}
		\label{pair_define}
	\end{subfigure}
	\begin{subfigure}[t]{0.4\textwidth}
		\centering
		\includegraphics[width=\textwidth]{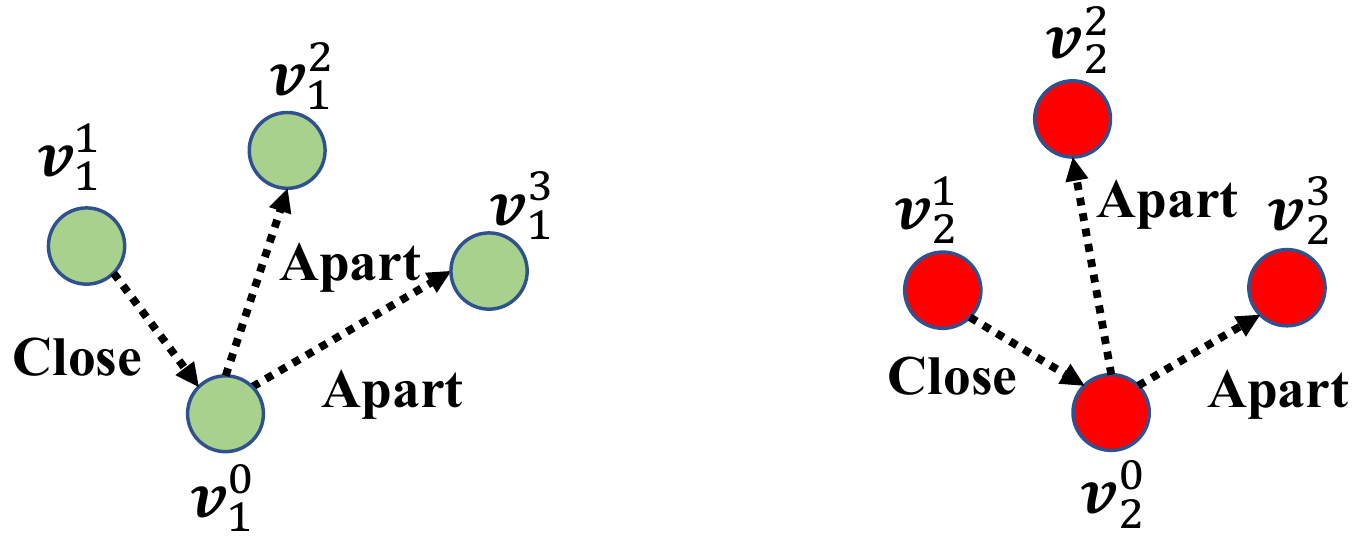}
		\caption{Vanilla Contrastive Learning (VCL).}
		\label{vcl}
	\end{subfigure}
	\begin{subfigure}[t]{0.43\textwidth}
		\centering
		\includegraphics[width=\textwidth]{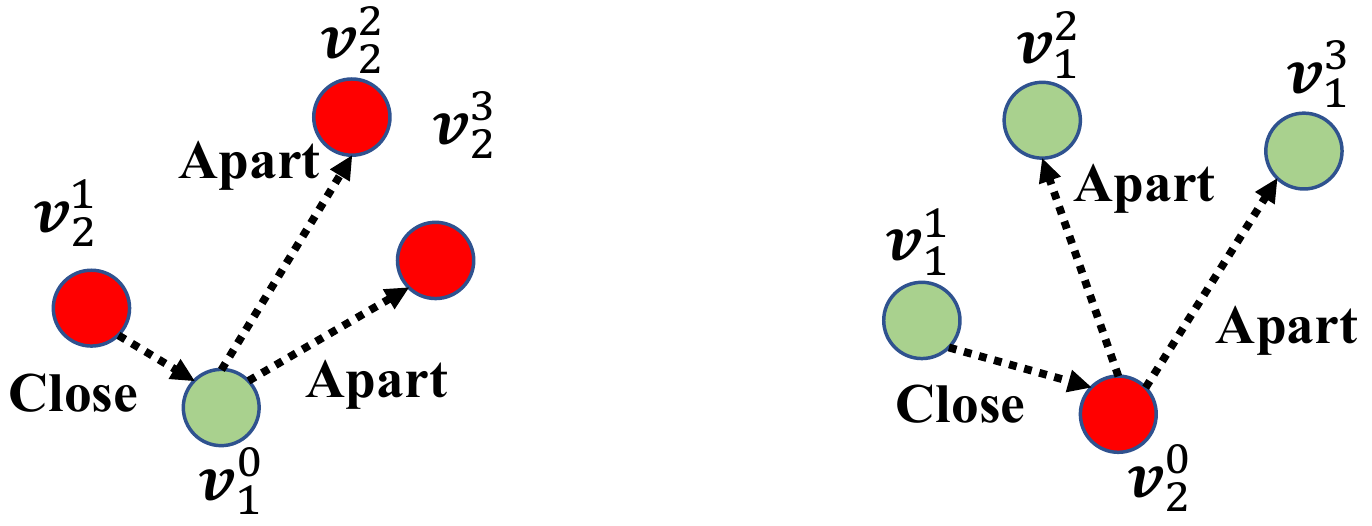}
		\caption{Interactive Contrastive Learning (ICL).}
		\label{icl_first}
	\end{subfigure}
	\caption{Overview of the proposed \emph{Mutual Contrastive Learning}. $f_{1}$ and $f_{2}$ denote two different networks. $\bm{v}_{m}^{i}$ is the embedding vector inferred from $f_{m}$ with the input sample $\bm{x}^{i}$. We use green and red colors to represent embeddings from $f_1$ and $f_2$, respectively. The dashed and dotted arrow denotes the direction we want to push close or apart by a contrastive loss. \textbf{The core difference between (b) VCL and (c) ICL is using contrastive embeddings from the same or different networks with the anchor. VCL uses the same color between an anchor and contrastive samples, while ICL uses different ones.}}
	\label{MCL_pre}
\end{figure} 
The main core of MCL is to perform mutual interaction and transfer of contrastive distributions among a cohort of models. MCL includes Vanilla Contrastive Learning (VCL) and Interactive Contrastive Learning (ICL). VCL follows the traditional contrastive paradigm that the positive and negative pairs are from the same network. Compared with the conventional VCL, our proposed ICL forms contrastive similarity distributions between diverse embedding spaces derived from two different networks. We demonstrate that the objective of ICL is equivalent to maximizing the lower bound to the mutual information between two peer networks. This can be understood to capture dependencies and enable a network to learn extra contrastive knowledge from another network.

Inspired by the idea of DML~\cite{zhang2018deep}, we also perform mutual alignment between different softmax-based contrastive distributions from various networks formed by the same data samples. Similar to DML~\cite{zhang2018deep}, the distributions can be seen as soft labels to supervise others. Such a peer-teaching manner with soft labels takes advantage of representation information embedded in different networks. Over two types of contrastive learning, we can derive \emph{soft VCL label} and \emph{soft ICL label}. Although the soft VCL label has been applied in previous KD works~\cite{ge2020mutual,fang2021seed}, its anchor and contrastive embeddings are still formed from the same network, limiting the information interactions. Instead, our proposed soft ICL label aggregates cross-network embeddings to construct contrastive distributions, which is demonstrated to be more informative than the conventional soft VCL label.  

To maximize the effectiveness of MCL, we summarize VCL and ICL with mutual mimicry into a unified framework, as illustrated in Fig.~\ref{MCL_pre}. MCL helps each model capture extra contrastive knowledge to construct a better representation space. As stated in DML~\cite{zhang2018deep}, since networks start from different initial conditions, each one can learn knowledge that others have not. MCL can be regarded as a group-wise contrastive loss over the feature level. Therefore, we can readily combine MCL with previous logit-level online KD methods. 

Beyond the final layer, we extend MCL to the intermediate layers between two networks, called \emph{layer-wise MCL}. The intermediate embeddings are refined by several attached feature modules. Moreover, we propose an adaptive layer-matching mechanism learned by meta-optimization, instead of the traditional manual matching. Empirical studies also show the superiority of our proposed layer-wise MCL. Besides, we construct a gated module to generate the weights of logit aggregation using all feature embeddings. The weights assemble logit distributions to a virtual peer teacher that performs online KD with other peers. 

We apply layer-wise MCL to image classification on CIFAR-100~\cite{krizhevsky2009learning} and ImageNet~\cite{deng2009imagenet} across various architectures. MCL can lead to consistent performance gains upon the baseline methods and outperforms other online KD methods. Extensive experiments on transfer learning to object detection and instance segmentation on COCO-2017~\cite{lin2014microsoft} show the superiority of layer-wise MCL to learn good features.  Note that collaborative learning among a cohort of models is conducted during the training stage. Any network in the cohort can be kept during the inference stage. 
The kept network does not introduce additional inference costs compared with the original network.

Parts of this paper were published originally in its conference version~\cite{yang2022mutual}. This paper extends our earlier work in several valuable aspects:
\begin{itemize}[noitemsep,nolistsep,,topsep=0pt,parsep=0pt,partopsep=0pt]
	\item We propose layer-wise MCL, an improved version by extending MCL to the intermediate features. We further propose an adaptive layer-matching mechanism learned by meta-optimization for layer-wise MCL.
	\item We introduce a logit-level online KD method by aggregating intermediate logit distributions using a gated module to construct a strong teacher role. This enables logits distillation from the virtual teacher to other peer networks.
	\item Beyond the same-style network pairs, we also evaluate layer-wise MCL over different architectures. Moreover, we further conduct transfer learning to downstream image classification, object detection and instance segmentation. The encouraging results are reported compared with competitive online KD methods, demonstrating that MCL can lead to better feature representations. 
	\item More detailed ablation studies and analyses are conducted to examine each component of layer-wise MCL.
\end{itemize}

\section{Related Work}

\subsection{Traditional Offline Knowledge Distillation} KD provides an elegant idea to improve a small student network supervised by a teacher network with better performance. The seminal Hinton's KD~\cite{hinton2015distilling} aims to guide the student to mimic the teacher's soft class predictions under the same data. This method is well-motivated since the student would learn a better predictive distribution. A convolutional neural network often encodes fruitful information beyond the final predictions. In recent years, many insightful approaches have been proposed to mining knowledge from the intermediate layers. Typical knowledge can be feature maps~\cite{romero2014fitnets,chen2021cross} and their extracted information, such as attention maps~\cite{zagoruyko2016paying}, activation boundaries~\cite{heo2019knowledge} and structural relations~\cite{park2019relational,peng2019correlation}, etc.

Apart from knowledge mining, another line aims to reduce the knowledge gap between the teacher and student. For example, some works~\cite{mirzadeh2020improved,passalis2020heterogeneous} introduce an assistant to explain the teacher's knowledge easily understandable to the student. RCO~\cite{jin2019knowledge} utilizes curriculum learning to supervise the student with those teachers from intermediate training states. However, a common drawback is that the offline KD follows a two-stage training pipeline. It requires us to pre-train a high-capacity teacher network and then perform KD from teacher to student, which is time-consuming. 

\subsection{Online Knowledge Distillation} Online KD is designed to enhance the student’s performance without a pre-trained teacher. The idea of online KD is to take advantage of collaborative learning among multiple student networks. DML~\cite{zhang2018deep} shows that a group of models can benefit from mutual learning of predictive class probability distributions from each other. CL~\cite{song2018collaborative} further extends this idea to a hierarchical architecture with multiple classifier heads. Over logit information, AFD~\cite{chung2020feature} introduces the mutual learning mechanism to feature maps using an adversarial training paradigm. The latest CKD-MKT~\cite{gou2022collaborative} guides multiple students to learn both individual instances and instance relations from each other through
collaborative and self-learning.


In contrast to mutual mimicry, some recent approaches, such as ONE~\cite{zhu2018knowledge}, OKDDip~\cite{chen2020online} and KDCL~\cite{guo2020online}, construct an online teacher via a weighted ensemble logit distribution but differ in various aggregation strategies. ONE~\cite{zhu2018knowledge} utilizes a learnable gate module to output aggregation weights. OKDDip~\cite{chen2020online} applies a self-attention mechanism to measure the similarities of group networks and improves peers' diversity to construct a better leader network. The group
leader is seen as the teacher role and used to transfer logits knowledge. KDCL~\cite{guo2020online} formulates aggregation weights as the optimal solution of the generalization error, solved by the Lagrange multiplier. More recently, PCL~\cite{wu2021peer} introduces an extra temporal mean network for each peer as the teacher role. This is because the temporal mean network is generated from averaged model weights, which is more accurate than the current model~\cite{tarvainen2017mean}. FFSD~\cite{li2022distilling} fuses feature maps
from auxiliary students to boost the leader student. A more comprehensive survey of online KD approaches can be referred to~\cite{gou2021knowledge}.

We can conclude that previous approaches often perform logit-level distillation among multiple networks and mainly differ in various learning mechanisms. Beyond the logit level, our method takes advantage of collaborative learning from the perspective of \emph{representation learning}. Moreover, we can readily incorporate MCL with previous logit-based methods together.

\subsection{Contrastive Learning} Contrastive learning has been extensively exploited for both supervised and self-supervised visual representation learning. The main idea of contrastive learning is to push positive pairs close and negative pairs apart by a contrastive loss~\cite{hadsell2006dimensionality} to \emph{obtain a discriminative space}. In the supervised scenario, labels often guide the definition of contrastive pairs. A positive pair is formed by two samples from the same class, while two samples from different classes form a negative pair.  Contrastive learning is widely used for deep metric learning, such as image classification~\cite{khosla2020supervised}, face recognition~\cite{schroff2015facenet}, person re-identification~\cite{hermans2017defense} and image retrieval~\cite{oh2016deep}. Recently, self-supervised contrastive learning can guide networks to learn pre-trained features for downstream visual recognition tasks. In the self-supervised scenario, since we do not have label information, a positive pair is often formed by two views of the same sample, while negative pairs are formed by different samples. Contrastive learning maximizes agreement between differently augmented views of the same data against negative samples to learn invariant representations~\cite{misra2020self,chen2020simple,he2020momentum} via a contrastive loss~\cite{oord2018representation}.  A more related work is SSKD~\cite{xu2020knowledge} which connects KD and contrastive learning. It introduces the idea of SimCLR~\cite{chen2020simple} to offline KD that guides the student to mimic a self-supervision pretext task from a teacher. Our focus is to take advantage of contrastive representation learning to improve online KD. MCL designs novel contrastive paradigms for multi-peer interaction, allowing each network to obtain better features from collaborative learning.

\subsection{Embedding-based Relational Distillation} Compared with the final class posterior, the latent feature embeddings encapsulates more structural information. Some previous KD methods transfer the embedding-based relational graph where each node represents one sample~\cite{park2019relational,peng2019correlation}. More recently, MMT~\cite{ge2020mutual} employs soft softmax-triplet loss to learn relative similarities from other networks for unsupervised domain adaptation on person Re-ID. To compress networks over self-supervised MoCo~\cite{he2020momentum}, SEED~\cite{fang2021seed} transfers soft InfoNCE-based~\cite{oord2018representation} contrastive distributions from a teacher to a student. A common characteristic of previous works is that contrastive distributions are often constructed from the same network, restricting peer information interactions. Instead, we aggregate cross-network embeddings to model interactive contrastive distributions and maximize the mutual information between two networks.

\section{Mutual Contrastive Learning}
\subsection{Architecture Formulation}
\textbf{Notation.} A classification network $f(\cdot)$ like ResNet~\cite{he2016deep} can be divided into a feature extractor $\varphi(\cdot)$ and a linear classifier $g(\cdot)$. $f$ maps an input image $\bm{x}$ to a logit vector $\bm{z}$, \emph{i.e.} $\bm{z}=f(\bm{x})=g(\varphi(\bm{x}))$.  Moreover, we add an additional projection module $\zeta(\cdot)$ that includes two sequential linear layers with a middle ReLU. $\zeta(\cdot)$ is to transform a feature embedding from the feature extractor $\varphi(\cdot)$ into a latent embedding $\bm{v}\in \mathbb{R}^{d}$, \emph{i.e.} $\bm{v}=\zeta(\varphi(\bm{x}))$, where $d$ is the embedding size. The embedding $\bm{v}$ is used for contrastive learning.

\begin{figure}[tbp]  
	\centering
	\includegraphics[width=1\linewidth]{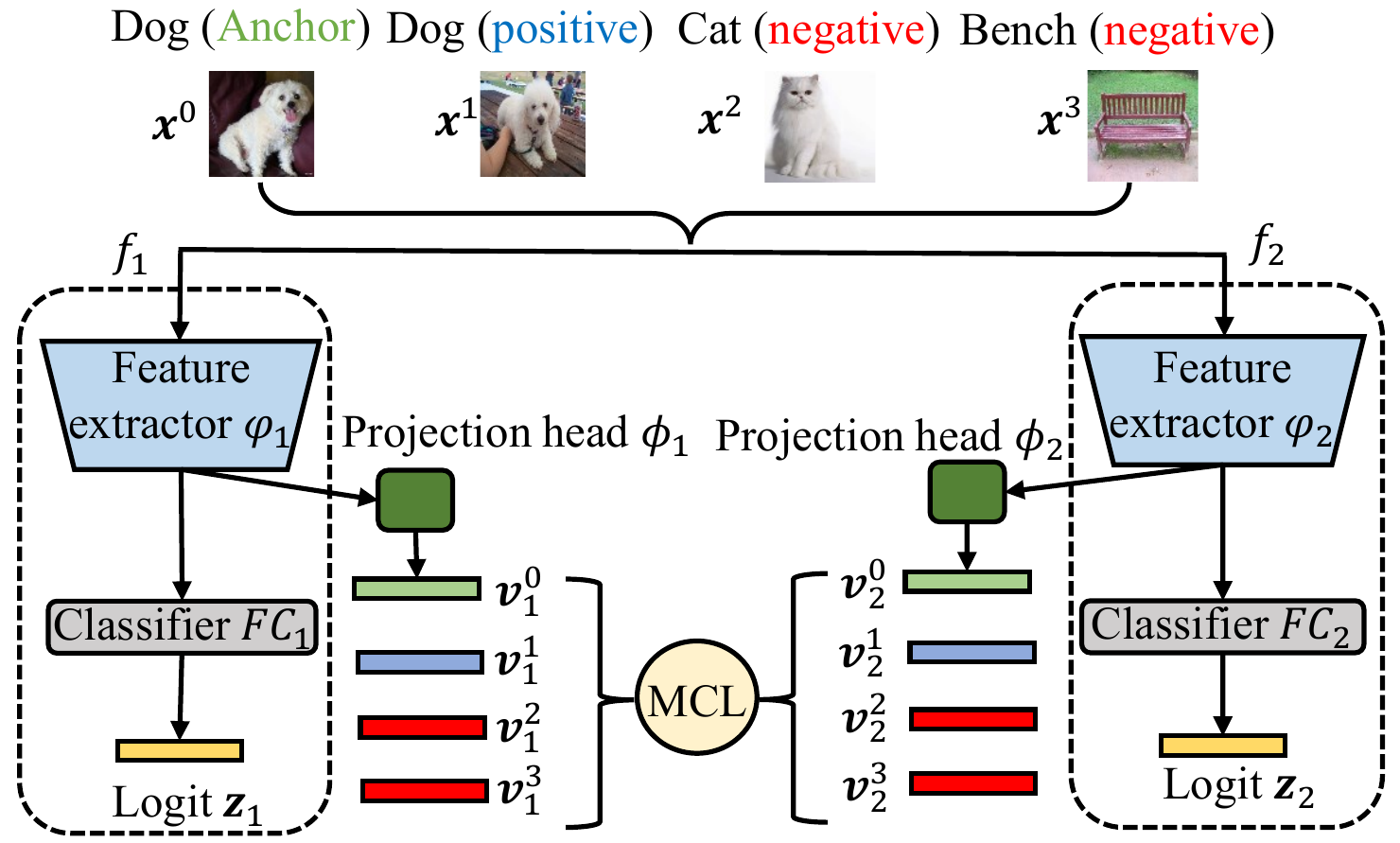}
	\caption{Overview of MCL between two networks of $f_{1}$ and $f_{2}$ modified from~\cite{yang2022mutual}.}
	\label{mcl_sup}
\end{figure}

\textbf{Training Graph.} The overall training graph contains $M (M\geqslant 2)$ classification networks denoted by $\{f_{m}\}_{m=1}^{M}$ for collaborative learning. All the same networks in the cohort are initialized with various weights to learn diverse representations. This is a prerequisite for the success of mutual learning. Each $f_{m}$ in the cohort is equipped with an additional embedding projection module $\zeta_{m}$. The overall training graph is shown in Fig.~\ref{mcl_sup}.

\textbf{Inference Graph.} During the test stage, we discard all projection modules $\{\zeta_{m}\}_{m=1}^{M}$ and keep one network for inference. We can select one network with the best validation performance in the cohort for final deployment. The architecture of the kept network is identical to the original network. That is to say that we do not introduce extra inference costs.

\subsection{Vanilla Contrastive Learning (VCL)}
Contrastive loss aims to push positive pairs close and negative pairs apart in the latent embedding space. Given an input sample $\bm{x}^{0}$ as the anchor, we retrieve $1$ positive sample $\bm{x}^{1}$ and $K(K\geqslant 1)$ negative samples $\{\bm{x}^{k}\}_{k=2}^{K+1}$. For supervised learning, the positive sample is from the same class with the anchor, while negative samples are from different classes. For ease of notation, we denote the anchor embedding as $\bm{v}_{m}^{0}$, the positive embedding as $\bm{v}_{m}^{1}$ and $K$ negative embeddings as $\{\bm{v}_{m}^{k}\}_{k=2}^{K+1}$. $m$ represents that the embedding is generated from $f_{m}$. Here, feature embeddings are preprocessed by $l_{2}$-normalization. 

We use the dot product to measure similarity distribution between the anchor and contrastive embeddings with \emph{softmax} normalization. Thus, we can obtain contrastive probability distribution:
\begin{equation}
\bm{p}_{m}=softmax([(\bm{v}_{m}^{0}\cdot \bm{v}_{m}^{1}/\tau),\cdots,(\bm{v}_{m}^{0}\cdot \bm{v}_{m}^{K+1}/\tau)]),
\end{equation}
where $\tau$ is a constant temperature. $\bm{p}_{m}\in \mathbb{R}^{K+1}$ measures the relative sample-wise similarities with a normalized probability distribution. A large probability represents a high similarity between the anchor and a contrastive embedding. We use cross-entropy loss to force the positive pair close and negative pairs away upon the contrastive distribution $\bm{p}_{m}$:
\begin{equation}
\mathcal{L}^{VCL}_{m}=-\log{\bm{p}_{m}^{1}}=-\log\frac{\exp(\bm{v}_{m}^{0}\cdot \bm{v}_{m}^{1}/\tau)}{\sum_{k=1}^{K+1}\exp (\bm{v}_{m}^{0}\cdot \bm{v}_{m}^{k}/\tau)}.
\label{scl}
\end{equation}
Here, $\bm{p}_{m}^{k}$ is the $k$-th element of $\bm{p}_{m}$. This loss is equivalent to a $(K+1)$-way softmax-based classification loss that forces the network to classify the positive sample correctly, also dubbed as an InfoNCE loss~\cite{oord2018representation}. When applying contrastive learning to a cohort of $M$ networks, the vanilla method is to summarize each contrastive loss:
\begin{equation}
\mathcal{L}^{VCL}_{1\sim M}=\sum_{m=1}^{M}(\mathcal{L}_{m}^{VCL}).
\end{equation}

\subsection{Interactive Contrastive Learning (ICL)}
However, vanilla contrastive learning does not model cross-network relationships for collaborative learning. This is because the contrastive distribution is learned from the network's own embedding space. To take full advantage of information interaction among various peer networks, we propose a novel \emph{Interactive Contrastive Learning} (ICL) to model cross-network interactions to learn better feature representations. We formulate ICL for the case of two parallel networks $f_{a}$ and $f_{b}$, where $a,b\in \{1,2,\cdots,M\},a\neq b$, and then further extend ICL to more than two networks among $\{f_{m}\}_{m=1}^{M}$.

To conduct ICL, we first fix $f_{a}$ and enumerate over $f_{b}$. Given the anchor embedding $\bm{v}_{a}^{0}$ extracted from $f_{a}$, we enumerate the positive embedding $\bm{v}_{b}^{1}$ and negative embeddings  $\{\bm{v}_{b}^{k}\}_{k=2}^{K+1}$ extracted from $f_{b}$. Here, both $\{\bm{v}_{a}^{k}\}_{k=0}^{K+1}$ and $\{\bm{v}_{b}^{k}\}_{k=0}^{K+1}$ are generated from the same $K+2$ samples $\{\bm{x}^{k}\}_{k=0}^{K+1}$ correspondingly, as illustrated in Fig.~\ref{pair_define}. The contrastive probability distribution from $f_{a}$ to $f_{b}$ can be formulated as:
\begin{equation}
\bm{q}_{a\rightarrow  b}=softmax([(\bm{v}_{a}^{0}\cdot \bm{v}_{b}^{1}/\tau),(\bm{v}_{a}^{0}\cdot \bm{v}_{b}^{2}/\tau),\cdots,(\bm{v}_{a}^{0}\cdot \bm{v}_{b}^{K+1}/\tau)]),
\end{equation}
where $\bm{q}_{a\rightarrow  b}\in \mathbb{R}^{K+1}$. Similar to Eq.(\ref{scl}), we use cross-entropy loss upon the contrastive distribution $\bm{q}_{a\rightarrow  b}$:
\begin{align}
\mathcal{L}^{ICL}_{a\rightarrow  b}&=-\log{\bm{q}_{a\rightarrow  b}^{1}}
=-\log\frac{\exp (\bm{v}_{a}^{0}\cdot \bm{v}_{b}^{1}/\tau)}{\sum_{k=1}^{K+1}\exp (\bm{v}_{a}^{0}\cdot \bm{v}_{b}^{k}/\tau)}.
\label{icl_hard}
\end{align}
Here, $\bm{q}_{a\rightarrow  b}^{k}$ is the $k$-th element of $\bm{q}_{a\rightarrow  b}$. We can observe that the main difference between Eq.(\ref{scl}) and Eq.(\ref{icl_hard}) lies in various types of embedding space for generating contrastive distributions. Compared with Eq.(\ref{scl}), Eq.(\ref{icl_hard}) employs contrastive embeddings from another network instead of the network's own embedding space. It can model explicit corrections or dependencies in various embedding spaces among multiple peer networks, facilitating  information communications to learn better feature representations.

Furthermore, compared to Eq.(\ref{scl}), we attribute the superiority of minimizing Eq.(\ref{icl_hard}) to maximizing the lower bound on the mutual information $I(\bm{v}_{a},\bm{v}_{b})$ between $f_{a}$ and $f_{b}$, which is formulated as:
\begin{equation}
I(\bm{v}_{a},\bm{v}_{b})\geq \log(K)- \mathbb{E}_{(\bm{v}_{a},\bm{v}_{b})}\mathcal{L}^{ICL}_{a\rightarrow  b}
\label{mutual_infor}.
\end{equation}

{The detailed proof from Eq.(\ref{icl_hard}) to derive Eq.(\ref{mutual_infor}) is provided in Appendix.} Intuitively, the mutual information $I(\bm{v}_{a},\bm{v}_{b})$ measures the reduction of uncertainty in contrastive feature embeddings from $f_{b}$ when the anchor embedding from $f_{a}$ is known. This can be understood that each network could gain extra contrastive knowledge from others benefiting from Eq.(\ref{icl_hard}). Thus, it can lead to better representation learning than independent contrastive learning of Eq.(\ref{scl}). As $K$ increases, the mutual information $I(\bm{v}_{a},\bm{v}_{b})$ would be higher, indicating that $f_{a}$ and $f_{b}$ could learn more common knowledge from each other. 

When extending to $\{f_{m}\}_{m=1}^{M}$, we perform ICL in every two of $M$ networks to model fully connected dependencies, leading to the overall loss as: 
\begin{equation}
\mathcal{L}_{1\sim M}^{ICL}=\sum_{1\leq a<b\leq M}^{M}(\mathcal{L}^{ICL}_{a\rightarrow  b}+\mathcal{L}^{ICL}_{b\rightarrow  a}) 
\end{equation}

\subsection{Soft Contrastive Learning with Online Mutual Mimicry}
The success of \emph{Deep Mutual Learning}~\cite{zhang2018deep} suggests that each network can generalize better from mutually learning other networks' soft class probability distributions in an online peer-teaching manner. This is because the output of class posterior from each network can be seen as a natural \emph{soft label} to supervise others. Based on this idea, it is desirable to derive soft contrastive distributions as \emph{soft labels} from contrastive learning, for example, $\bm{p}_{m}$ from VCL and $\bm{q}_{a\rightarrow  b}$ from ICL. In theory, both $\bm{p}_{m}$ and $\bm{q}_{a\rightarrow  b}$ can also be seen as class posteriors. Thus it is reasonable to perform mutual mimicry of these contrastive distributions for better representation learning.

Specifically, we utilize Kullback Leibler (KL)-divergence to force each network's contrastive distributions to align corresponding soft labels provided from other networks within the cohort. This paper focuses on mutually mimicking two types of contrastive distributions from VCL and ICL:

\subsubsection{Soft Vanilla Contrastive Learning (Soft VCL)} For refining $\bm{p}_{m}$ from $f_{m}$, the soft pseudo labels are peer contrastive distributions $\{\bm{p}_{l}\}_{l=1,l\neq m }^{l=M}$ generated from $\{f_{l}\}_{l=1,l\neq m }^{l=M}$, respectively. We use $\mathbf{KL}$ divergence to force $\bm{p}_{m}$ to align them. For applying soft VCL to the cohort of $\{f_{m}\}_{m=1}^{M}$, the overall loss can be formulated as:
\begin{equation}
\mathcal{L}_{1\sim M}^{Soft\_VCL}=\sum_{m=1}^{M}\sum_{l=1,l\neq m }^{M}\mathbf{KL
}(\bm{p}_{l}\parallel \bm{p}_{m}).
\label{VSCL}
\end{equation}
Here, $\bm{p}_{l}$ is the soft label detached from gradient back-propagation for stability.

\subsubsection{Soft Interactive Contrastive Learning (Soft ICL)} Given two networks $f_{a}$ and $f_{b}$, we can derive interactive contrastive distributions $\bm{q}_{a\rightarrow b}$ and $ \bm{q}_{b\rightarrow a}$ using ICL. It makes sense to force the consistency between $\bm{q}_{a\rightarrow b}$ and $ \bm{q}_{b\rightarrow a}$ for mutual calibration by Soft ICL. When extending to $\{f_{m}\}_{m=1}^{M}$, we perform Soft ICL in every two of $M$ networks, leading to the overall loss as: 
\begin{equation}
\mathcal{L}_{1\sim M}^{Soft\_ICL}=\sum_{a=1}^{M}\sum_{b=1,b\neq a }^{M}\mathbf{KL
}(\bm{q}_{b\rightarrow a}\parallel \bm{q}_{a\rightarrow b}).
\label{VICL}
\end{equation}
Here, $\bm{q}_{b\rightarrow a}$ is the soft label detached from gradient back-propagation for stability.

\subsubsection{Discussion with Soft VCL and Soft ICL} We remark that using a vanilla contrastive distribution $\bm{p}$ as a soft label has been explored by some previous works~\cite{ge2020mutual,fang2021seed}. These works often construct contrastive relationships using embeddings from the same network, as illustrated in Eq.(\ref{scl}). In contrast, we propose an interactive contrastive distribution $\bm{q}$ to perform Soft ICL. Intuitively, $\bm{q}$ aggregates cross-network embeddings to model the soft label, which is more informative than $\bm{p}$ constructed from a single embedding space. Moreover, refining a better $\bm{q}$ may decrease $\mathcal{L}^{ICL}_{a\rightarrow  b}$, further maximizing the lower bound on the mutual information $I(\bm{v}_{a}, \bm{v}_{b})$ between $f_{a}$ and $f_{b}$. Compared with soft VCL, soft ICL can facilitate more adequate interactions among multiple networks. Empirically, we found soft ICL excavates better performance gains by taking full advantage of collaborative contrastive learning, as verified in Section~\ref{Ablation}. 

\subsection{Overall loss of MCL}
To take full advantage of collaborative learning, we summarize all contrastive loss terms as the overall loss for MCL among a cohort of $M$ networks:

\begin{align}
\mathcal{L}^{MCL}_{1\sim M}(\{\bm{v}_{m}\}_{m=1}^{M})=&\alpha(\mathcal{L}^{VCL}_{1\sim M}+\mathcal{L}^{ICL}_{1\sim M}) \notag \\
&+\beta(\mathcal{L}^{Soft\_VCL}_{1\sim M}+\mathcal{L}^{Soft\_ICL}_{1\sim M}),
\label{MCL}
\end{align}

where $\alpha$ and $\beta$ are weight coefficients. We set $\alpha=0.1$ for hard cross-entropy losses and $\beta=1$ for soft KL-divergence losses, as suggested by our empirical study in Section~\ref{Ablation}.

\section{Layer-wise Mutual Contrastive Learning}

The original MCL applies contrastive learning to those feature embeddings from the final layer. This ignores the intermediate features that represent the abstraction process of the input image. To further improve representation learning, we propose \textbf{\emph{Layer-wise Mutual Contrastive Learning}} (L-MCL). It further excavates MCL by allowing the representation learning from the intermediate layers.

\subsection{Architecture Formulation}
\textbf{Training Graph.}  The overall training graph contains $M (M\geqslant 2)$ classification networks denoted by $\{f_{m}\}_{m=1}^{M}$ with augmented modules for collaborative learning, as shown in Appendix. The $m$-th classification network $f_{m}(\cdot)$ like ResNet~\cite{he2016deep} can be divided into a feature extractor $\varphi_{m}(\cdot)$ and a linear classifier $g_{m}(\cdot)$, where $\varphi_{m}=\varphi_{m}^{[L]}\circ\varphi_{m}^{[L-1]}\cdots\varphi_{m}^{[1]}$ and $L$ is the number of stages in $\varphi_{m}$. After each intermediate stage $\varphi^{[l]}_{m}$, where $l=1,2,\cdots,L-1$, we attach an auxiliary feature refinement module $r_{m}^{[l]}(\cdot)$, a projection head $\zeta_{m}^{[l]}(\cdot)$ for contrastive learning and a linear classifier $g_{m}^{[l]}(\cdot)$ for learning ground-truth labels. The feature refinement module $r_{m}^{[l]}(\cdot)$ includes block-wise convolutional modules and global average pooling to output embeddings. For easy notation, we denote the feature of the $l$-th stage as $F_{m}^{[l]}$, $l=1,2,\cdots,L$:
\begin{align}
&F_{m}^{[1]}(\bm{x})=r_{m}^{[1]}\circ \phi_{m}^{[1]}(\bm{x}), \notag \\
&F_{m}^{[2]}(\bm{x})=r_{m}^{[2]}\circ \phi_{m}^{[2]}\circ \phi_{m}^{[1]}(\bm{x}), \notag \\
&\cdots \notag \\
&F_{m}^{[L-1]}(\bm{x})=r_{m}^{[L-1]}\circ  \phi_{m}^{[L-1]}\circ \cdots \circ \phi_{m}^{[1]}(\bm{x}), \notag\\
& F_{m}^{[L]}(\bm{x})=\phi_{m}(\bm{x}).
\end{align}

Therefore, given an input image $\bm{x}$, we have $L$ contrastive embeddings $\bm{v}_{m}^{[1]},\cdots,\bm{v}_{m}^{[L]}$ and $L$ class logits $\bm{z}_{m}^{[1]},\cdots,\bm{z}_{m}^{[L]}$ that are formulated as:
\begin{equation}
\bm{v}^{[l]}_{m}=\zeta_{m}^{[l]}(F^{[l]}_{m}(\bm{x})),\ \bm{z}^{[l]}_{m}=g_{m}^{[l]}(F_{m}^{[l]}(\bm{x})).
\end{equation}
Here, $l=1,2,\cdots,L$. We denote $g_{m}^{[L]}=g_{m}$ and $\zeta_{m}^{[L]}=\zeta_{m}$ as the classifier and projection head after the final layer.

\textbf{Inference Graph.} During the test stage, we discard all auxiliary components and keep one network for inference. We can select one network with the best validation performance in the cohort for final deployment. The architecture of the kept network is identical to the original network without extra inference costs. 

\subsection{Layer-wise MCL}
\begin{figure}[t]
	\centering 
	
	\begin{subfigure}[t]{0.24\textwidth}
		\centering
		\includegraphics[width=\textwidth]{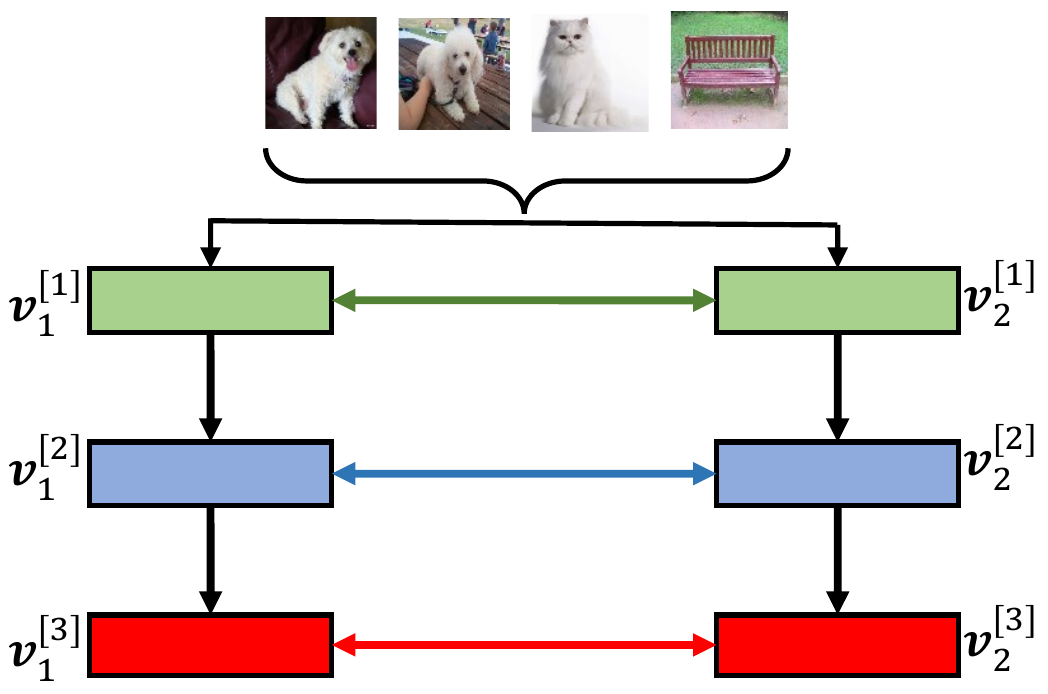}
		\caption{One-to-one match.}
		\label{one_to_one}
	\end{subfigure}
	\begin{subfigure}[t]{0.24\textwidth}
		\centering
		\includegraphics[width=\textwidth]{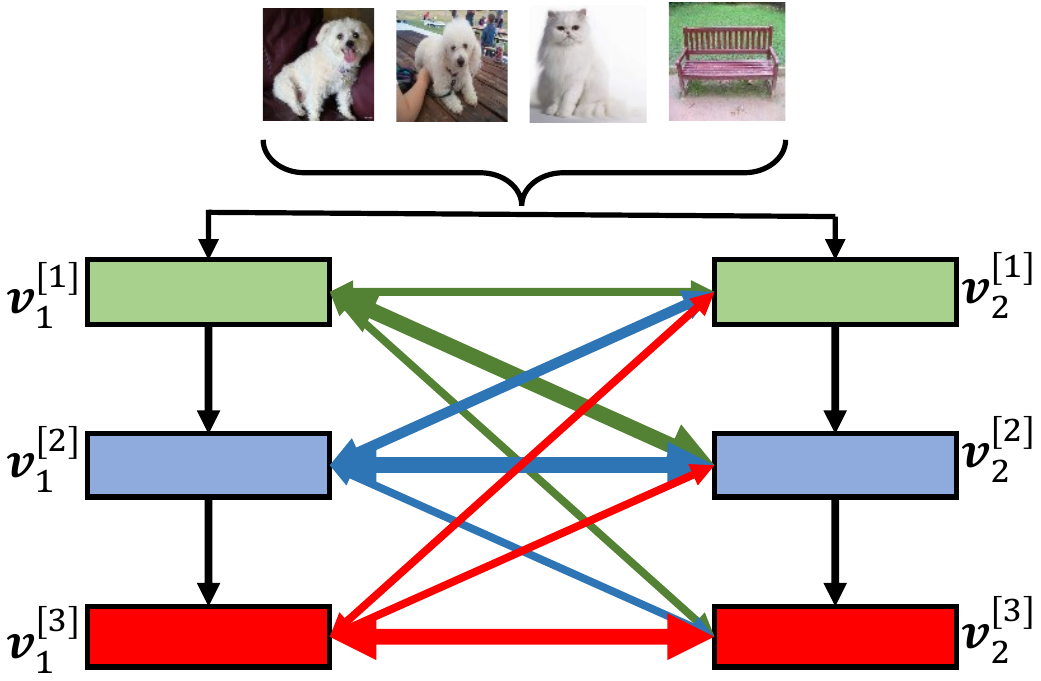}
		\caption{Weighted all-to-all match.}
		\label{all_match}
	\end{subfigure}
	\caption{Overview of the proposed one-to-one match and weighted all-to-all match. The two-way arrow represents the layer association.}
	\label{match}
\end{figure} 
Beyond the final layer, we also aim to conduct MCL over the intermediate layers between peer networks. Therefore, the layer-matching algorithm becomes a critical problem to be considered. As we all know, some conventional KD methods~\cite{romero2014fitnets,zagoruyko2016paying} often use vanilla one-to-one matching between the same-staged intermediate feature layers. However, the intermediate layers of peer networks may have distinct semantic levels~\cite{passalis2020heterogeneous}, especially on different network architectures. A more reasonable practice is to perform adaptive layer association for intermediate feature matching using a weighted mechanism. Therefore, a feature layer could select which feature layers from another network are more useful for its own contrastive learning. The detailed formulation of layer-wise MCL are shown as follows.

Given the cohort $\{f_{m}\}_{m=1}^{M}$ with $L$ stages in each network, the original MCL deals with the final feature embeddings of $\{\bm{v}^{[L]}_{m}\}_{m=1}^{M}$. Layer-wise MCL further extends mutual contrastive learning to intermediate and final embeddings $\{\{\bm{v}^{[l]}_{m}\}_{m=1}^{M}\}_{l=1}^{L}$ in a cross-layer manner:
\begin{align}
&\mathcal{L}^{L\_MCL}_{1\sim M}(\theta |\bm{x},y,\pi) \notag\\
&=\sum_{a=1}^{M}\sum_{b=1,b\neq a}^{M}\sum_{la=1}^{L}\sum_{lb=1}^{L}\lambda_{a,b}^{la,lb}\mathcal{L}^{MCL}_{a,b}(\bm{v}^{[la]}_{a},\bm{v}^{[lb]}_{b}),
\label{L_MCL}
\end{align}
where $\mathcal{L}^{MCL}_{a,b}(\bm{v}^{[la]}_{a},\bm{v}^{[lb]}_{a})$ is the original MCL loss between the $f_{a}$ and $f_{b}$ networks with regard to the feature embeddings of $\bm{v}^{[la]}_{a}$ and $\bm{v}^{[lb]}_{b}$. $\bm{v}^{[la]}_{a}$ is the $la$-th layer's embedding from the network $f_{a}$, and $\bm{v}^{[lb]}_{b}$ is the $lb$-th layer's embedding from the network $f_{b}$. $\lambda_{a,b}^{la,lb}\in (0,1)$ is the learnable matching weight parameter between the $la$-th layer of network $f_{a}$ and the $lb$-th layer of network $f_{b}$. Here, for ease notation, we denote the parameters of $\{f_{m}\}_{m=1}^{M}$ as $\theta$, the input sample as $\bm{x}$ with the label $y$. In the next section, we show how to optimize $\lambda$ using meta-networks parameterized by $\pi$.

\subsection{Training Meta-Networks}
\textbf{Basic cross-entropy task loss}. For the supervised image classification, we train $M$ networks $\{f_{m}\}_{m=1}^{M}$ with auxiliary components via the conventional cross-entropy loss $\mathcal{L}_{ce}$ with the groud-truth label $y$ as the basic task loss:
\begin{equation}
\mathcal{L}^{task}_{1\sim M}(\theta|\bm{x},y)=\sum_{m=1}^{M}\sum_{l=1}^{L}\mathcal{L}_{ce}(\sigma(\bm{z}_{m}^{[l]}),y)
\label{task_loss}
\end{equation}
Here, $\sigma$ denotes the softmax function. The task loss, a prerequisite of mutual learning, guides the networks to learn task-aware information. We summarize the basic task loss and layer-wise MCL loss as the total loss for feature-level online KD:
\begin{equation}
\mathcal{L}^{total}_{1\sim M}(\theta|\bm{x},y,\pi)=\mathcal{L}^{task}_{1\sim M}(\theta|\bm{x},y)+\mathcal{L}^{L\_MCL}_{1\sim M}(\theta |\bm{x},y,\pi).
\end{equation}

\textbf{Meta-optimization.} Our final objective is to achieve excellent classification performance using the training loss $\mathcal{L}^{total}_{1\sim M}(\theta|\bm{x},y,\pi)$. To realize this goal, the core layer-wise MCL loss of $\mathcal{L}^{L\_MCL}_{1\sim M}(\theta |\bm{x},y,\pi)$ should let networks learn good feature representations, benefiting the classification. To measure and improve the feature matching capabilities guided by the meta-network $\pi$, a desirable method is to train it with a bilevel scheme~\cite{colson2007overview}:

\begin{enumerate}
	\item Update $\theta$ to minimize $\mathcal{L}^{total}_{1\sim M}(\theta|\bm{x},y,\pi)$ for $K$ times.
	\item Measure $\mathcal{L}^{task}_{1\sim M}(\theta|\bm{x},y)$ and update $\pi$ to minimize it.
\end{enumerate}
First, we optimize $\theta$ by minimizing the total loss given the meta-network $\pi$. Then  the basic task loss $\mathcal{L}^{task}_{1\sim M}(\theta|\bm{x},y)$ is regarded as a meta-objective to measure the effectiveness of the meta-network $\pi$ for weighted layer association. However, we find that the meta-network $\pi$ impacts the training procedure through $\mathcal{L}^{L\_MCL}$ slower than the original task loss $\mathcal{L}^{task}$. It may result in insufficient optimization of $\pi$ using the
gradient $\nabla_{\pi}\mathcal{L}^{task}$. To address this problem, we conduct an alternative scheme:
\begin{enumerate}
	\item Update $\theta$ to minimize $\mathcal{L}^{L\_MCL}_{1\sim M}(\theta |\bm{x},y,\pi)$ for $K$ times.
	\item Update $\theta$ to minimize $\mathcal{L}^{task}_{1\sim M}(\theta|\bm{x},y)$ once.
	\item Measure $\mathcal{L}^{task}_{1\sim M}(\theta|\bm{x},y)$ and update $\pi$ to minimize it.
\end{enumerate}
At the first stage, we optimize the initial parameter $\theta_0$ for $K$ times by minimizing $\mathcal{L}^{L\_MCL}$. The obtained $\theta_{K}$ is learned only from the feature-based layer-wise MCL. It emphasizes the impact of weighted layer-matching along the training procedure of the networks. Here, we find $K=2$ is good enough. The second stage is a one-step adaptation from $\theta_{K}$ to $\theta_{K+1}$ by minimizing the basic task loss. At the third stage, the task-oriented loss $\mathcal{L}^{task}(\theta_{K+1})$ evaluates how quickly the networks have adapted to the task  via one step from $\theta_{K}$ to $\theta_{K+1}$ through data samples utilized in the first and second stages. Finally, the meta-network $\pi$ is optimized by minimizing $\mathcal{L}^{task}(\theta_{K+1})$. This three-stage mechanism allows more direct and faster optimization of meta-network $\pi$ from $\mathcal{L}^{L\_MCL}$ solely than the standard two-stage procedure. In summary, the optimization objective of the meta-network $\pi$ is formulated as:
\begin{align}
	&\min_{\pi} \mathcal{L}^{task}_{1\sim M}(\theta_{K+1}|\bm{x},y) \label{three_stage}\\
	 s.t.\ &\theta_{K+1}=\theta_{K}-\eta \nabla_{\theta}\mathcal{L}^{task}_{1\sim M}(\theta_{K}|\bm{x},y), \notag\\
	 &\theta_{k+1}=\theta_{k}-\eta \nabla_{\theta}\mathcal{L}^{L\_MCL}_{1\sim M}(\theta_{k} |\bm{x},y,\pi), \notag\\
	 &k=0,1,\cdots,K-1, \notag
\end{align}
where $\eta$ denotes the learning rate. We further adopt  Reverse-HG~\cite{franceschi2017forward} to solve the above optimization. It utilizes  Hessian-vector products to calculate $\nabla_{\pi}\mathcal{L}^{task}_{1\sim M}(\theta_{K+1}|\bm{x},y)$ effectively.

To learn weighted layer-matching via meta-optimization, we alternatively update the peer networks' parameters $\theta$ and the meta-network's parameters $\pi$. For the training scheme, it first updates $\theta$ by minimizing $\mathcal{L}^{total}_{1\sim M}(\theta|\bm{x},y,\pi)$, then updates $\pi$ through the three-stage bilevel optimization of Equ(\ref{three_stage}). We summarize the training procedure of meta-optimization in Algorithm~\ref{meta_opt}. To save training costs in practice, we perform meta-optimization every several mini-batch iterations.

\begin{algorithm}[tb]
	\caption{Meta optimization of parameters $\theta$ and $\pi$}
	\begin{algorithmic}
		\label{meta_opt}
		\WHILE{ $\theta$ and $\pi$ have not converged}
		\STATE 	Sample a mini-batch $\{\bm{x}^{(i)},{y}^{(i)}\}_{i=1}^{B}$, $B$ is batch size, $\bm{x}^{(i)}$ is the $i$-th sample and ${y}^{(i)}$ is the ground-truth label.
		Update $\theta$ by minimizing $\frac{1}{B}\sum_{i=1}^{B}\mathcal{L}^{total}_{1\sim M}(\theta|\bm{x}^{(i)},y^{(i)},\pi)$
		\STATE Initialize $\theta_0 \gets \theta$
		\FOR{$k = 0 \to K-1$}
		\STATE $\theta_{k+1}\gets \theta_{k}-\eta \nabla_{\theta}\frac{1}{B}\sum_{i=1}^{B}\mathcal{L}^{L\_MCL}_{1\sim M}(\theta_{k} |\bm{x}^{(i)},y^{(i)},\pi)$
		\ENDFOR
		\STATE $\theta_{K+1}\gets \theta_{K}-\eta \nabla_{\theta}\frac{1}{B}\sum_{i=1}^{B}\mathcal{L}^{task}_{1\sim M}(\theta_{K}|\bm{x}^{(i)},y^{(i)})$
		\STATE Update $\pi$ using $\nabla_{\pi}\frac{1}{B}\sum_{i=1}^{B}\mathcal{L}^{task}_{1\sim M}(\theta_{K+1}|\bm{x}^{(i)},y^{(i)})$
		\ENDWHILE
	\end{algorithmic}
\end{algorithm}

\textbf{Meta-network $\pi$ architecture.} The meta-network includes two linear projection layers $\xi^{la}_{a}\in \mathbb{R}^{d\times d}$ and $\xi^{lb}_{b}\in \mathbb{R}^{d\times d}$ for each feature embedding pairs of $(\bm{v}^{[la]}_{a}\in \mathbb{R}^{1\times d},\bm{v}^{[lb]}_{b}\in \mathbb{R}^{1\times d})$, where $a,b\in [1,2,\cdots,M]$, $la,lb\in [1,2,\cdots,L]$. After projection, the embeddings are preprocessed by a $l_2$ normalization operator $Norm(\cdot)$. Inspired by the self-attention mechanism~\cite{zhang2016colorful}, we adopt feature similarity via dot product to measure the correlation of matched layers. We further introduce a sigmoid activation function $\delta (\cdot)$ to scale the similarity value to $(0,1)$ as the layer-matching weight $\lambda_{a,b}^{la,lb}$. The procedure is formulated as:
\begin{equation}
	\lambda_{a,b}^{la,lb}=\delta(Norm(\xi^{la}_{a}(\bm{v}^{[la]}_{a}))Norm(\xi^{lb}_{b}(\bm{v}^{[lb]}_{b}))^\top)
\end{equation}

\subsection{Logit-level Online Knowledge Distillation}

Beyond the feature level, previous online KD methods often conduct logit-level collaborative learning among a group of networks. The class probability distribution has been widely demonstrated as an informative form of mutual learning. We further propose a simple yet effective ensemble distillation over the training architecture of layer-wise MCL. For each network $f_{m}$, we aggregate multi-branch logits $\{\bm{z}^{[l]}_{m}\}_{l=1}^{L}$ via fused weights of $\{w_{m}^{[l]}\}_{l=1}^{L}$ to build a virtual teacher role denoted as $\bm{z}^{ens}_{m}$:
\begin{equation}
\bm{z}^{ens}_{m}=\sum_{l=1}^{L}w_{m}^{[l]}\cdot \bm{z}^{[l]}_{m},
\label{w_ens}
\end{equation}
where $w_{m}^{[l]}$ is the importance weight of the $l$-th layer's logits $\bm{z}^{[l]}_{m}$. We create a learnable gate module $G_{m}$ to produce the aggregation weights of $\bm{w}$ according to the channel concatenation of input feature embeddings $\{F_{m}^{[l]}\}_{l=1}^{L}$ as the input clues:
\begin{equation}
\bm{w}_{m}=G_{m}(concat(\{F_{m}^{[l]}\}_{l=1}^{L})),
\end{equation}
where $\bm{w}_{m}=[w_{m}^{[1]},w_{m}^{[2]},\cdots,w_{m}^{[L]}]$. The gate module $G_{m}$ comprises two linear layers with a middle ReLU activation and terminated by softmax, where the last layer with $L$ neurons is to output weight values. $\bm{z}^{ens}_{m}$ is also supervised by the label $y$ via cross-entropy loss $\mathcal{L}_{ce}$ to make $G_{m}$ learnable:
\begin{equation}
\mathcal{L}^{task\_G}_{1\sim M}=\sum_{m=1}^{M}\mathcal{L}_{ce}(\sigma(\bm{z}_{m}^{ens}),y).
\end{equation}

Given the cohort $\{f_{m}\}_{m=1}^{M}$, we can obtain the corresponding $M$ ensemble logits $\{\bm{z}^{ens}_{m}\}_{m=1}^{M}$. Then we perform mutual learning among $M$ networks in a traditional peer-teaching manner via KL-divergence. Given the $a$-th network's final logit $\bm{z}^{[L]}_{a}$, the other $M-1$ networks' ensemble logits $\{\bm{z}^{ens}_{b}\}_{b=1,b\ne a}^{M}$ are regarded as teachers to supervise $f_{a}$:
\begin{equation}
\mathcal{L}^{ens}_{1\sim M}=T^{2}\sum_{a=1}^{M}\sum_{b=1,b\ne a}^{M}\mathbf{KL
}(\sigma(\frac{\bm{z}^{ens}_{b}}{T})\parallel \sigma(\frac{\bm{z}^{[L]}_{a}}{T})),
\end{equation} 
where $T$ is a temperature constant and we set $T=3$ following~\cite{zhu2018knowledge}. As suggested by~\cite{miyato2018virtual}, the gradient is not propagated through the teacher distribution $\bm{z}^{ens}_{b}$ to avoid the model collapse problem.

We summarize the gate module task loss $\mathcal{L}^{task\_G}_{1\sim M}$ and ensemble-based distillation loss $\mathcal{L}^{ens}_{1\sim M}$ for joint optimization, leading to the logit-based distillation loss as:
\begin{equation}
	\mathcal{L}^{logit}_{1\sim M}=\mathcal{L}^{task\_G}_{1\sim M}+\mathcal{L}^{ens}_{1\sim M}.
	\label{logit_loss}
\end{equation}

\subsection{Training and Optimization}
\label{Optimization}

\subsubsection{Overall loss}
We summarize the task loss $\mathcal{L}^{task}_{1\sim M}$ (Eq.(\ref{task_loss})) , logit-based distillation loss $\mathcal{L}^{logit}_{1\sim M}$ (Eq.(\ref{logit_loss}))  and embedding-based layer-wise MCL loss $\mathcal{L}^{L\_MCL}_{1\sim M}$ (Eq.(\ref{L_MCL})) for collaborative learning:
\begin{equation}
\label{overall_loss}
\mathcal{L}=\mathcal{L}^{task}_{1\sim M}+\mathcal{L}^{logit}_{1\sim M}+\mathcal{L}^{L\_MCL}_{1\sim M}.
\end{equation}
We do not introduce weights for probability-based cross-entropy and KL-divergence losses since these entropy-based losses often have the same magnitude. We illustrate the overview of the training framework in Appendix Fig. 1.

\subsubsection{Contrastive sample mining} One critical problem in contrastive learning is contrastive sample mining. We consider two implementations as follows:

\textbf{(1) Batch-based mining.} We create a class-aware sampler to construct mini-batches. The mini-batch with a batch size of $B$ consists of $B/2$ classes. Each class has two samples, and others from different classes are negative samples. We regard each sample as an anchor instance and others as contrastive instances within the current mini-batch.

\textbf{(2) Memory-based mining.} The batch size limits the number of available contrastive samples. Wang \emph{et al.}~\cite{wang2020cross} show that feature embeddings drift slowly throughout the training process. We create an online memory bank~\cite{wu2018unsupervised} during the training to store massive embeddings from past iterations. This allows us to retrieve sufficient contrastive samples from the memory bank for each training step. 

According to our empirical study in Section~\ref{Ablation}, we utilize batch-based mining for the small-scale dataset like CIFAR-100~\cite{krizhevsky2009learning} and memory-based mining for the large-scale dataset like ImageNet~\cite{deng2009imagenet}. \emph{The pseudo-code of the implementation details of MCL on CIFAR-100 and ImageNet with various mining approaches is shown in Appendix.}

\subsubsection{Training complexity}
We examine the training complexity of layer-wise MCL. Consistent with the above, we denote the embedding dimension as $d$ and the number of negative samples as $K$. The first step is to compute contrastive distributions of $\{\bm{p}_{a\rightarrow  b}\}_{a=1,b=1}^{M}$ among the $M$ networks, whose computational complexity is $\mathcal{O}(M^2Kd)$. Cross-entropy-based VCL and ICL minimize negative log-likelihood, whose complexity is negligible given the contrastive distribution. The second step is to perform mutual mimicry among $\{\bm{p}_{a\rightarrow  b}\}_{a=1,b=1}^{M}$ via soft VL, whose computational complexity is $\mathcal{O}(M^2K)$. Since we conduct layer-to-layer MCL among networks, the total computational complexity is  $L^{2}(\mathcal{O}(M^2Kd)+\mathcal{O}(M^2K))=\mathcal{O}(M^2 L^2 Kd)$, where $L$ is the number of network layers. When $L=1$, the complexity changes to $\mathcal{O}(M^2 Kd)$, corresponding to the original MCL. For logit-level online KD, the class probability distribution of each network is guided to learn from other $M-1$ networks, leading to the computational complexity of $\mathcal{O}(M^2C)$, where $C$ is the number of classes. As a result, the overall computational complexity of our proposed approach is $\mathcal{O}(M^2(L^2 Kd+C))$.  Compared with other online KD methods over the logit level, our approach mainly takes extra computational costs for feature-level contrastive learning. 

Empirically, we take two ResNet-50 networks trained for ImageNet classification as an example. L-MCL takes extra 16 MFLOPs for layer-to-layer mutual contrastive learning, around 0.2\% of the original 8 GFLOPs for baseline training. MCL needs extra 1MFLOPs for contrastive learning. In practice, the actual training time is measured on 8 NVIDIA Tesla A100 GPUs.  For independently training two baseline networks, the time is about 14min/epoch and the per GPU memory is about 12GB. We also evaluate the training costs of the original MCL and SOTA CKD-MKT. The former takes 16min/epoch and 13.7GB/GPU, and the latter takes 17min/epoch and 12.5GB/GPU. In contrast, our L-MCL consumes 20min/epoch and 15.8GB/GPU. Overall, L-MCL does not introduce much training cost but achieves significant performance improvements.

\section{Experiments}
In this section, we evaluate the proposed layer-wise MCL (called \emph{L-MCL} for simplicity) on CIFAR-100~\cite{krizhevsky2009learning} and ImageNet~\cite{deng2009imagenet} classification tasks compared with state-of-the-art online KD methods across various network pairs. Extensive transfer experiments to image classification, detection and segmentation are conducted to examine the effectiveness of learned feature representations. Finally, we show detailed ablation studies and parameter analyses to investigate each component. 

\begin{table*}[t]
	\centering 
	\caption{Top-1 accuracy (\%) of online KD methods by jointly training \textbf{two networks with the same architecture} on CIFAR-100. The \textbf{bold number} is the best result among various methods, while the \underline{underline number} is the second best.}
	\resizebox{1.\linewidth}{!}{
		\begin{tabular}{c|ccccccc|cc}  
			\hline
			Network&Baseline&DML~\cite{zhang2018deep}&ONE~\cite{zhu2018knowledge}&AFD~\cite{chung2020feature}&KDCL~\cite{guo2020online}&PCL~\cite{wu2021peer}&CKD-MKT~\cite{gou2022collaborative}&MCL~\cite{yang2022mutual}&L-MCL(Ours) \\  
			\hline
			ResNet-32~\cite{he2016deep}&70.91$_{\pm 0.14}$&72.14$_{\pm 0.35}$&73.71$_{\pm 0.25}$&72.15$_{\pm 0.13}$&74.22$_{\pm 0.27}$&\underline{74.52}$_{\pm 0.42}$&72.84$_{\pm 0.53}$&73.84$_{\pm 0.12}$&\textbf{75.82}$_{\pm 0.24}$\\
			
			ResNet-56~\cite{he2016deep}&73.15$_{\pm 0.23}$&75.36$_{\pm 0.12}$&75.46$_{\pm 0.27}$&75.72$_{\pm 0.13}$&75.10$_{\pm 0.36}$&\underline{76.22}$_{\pm 0.38}$&76.13$_{\pm 0.28}$&75.88$_{\pm 0.23}$&\textbf{77.51}$_{\pm 0.28}$ \\
			
			ResNet-110~\cite{he2016deep}&75.29$_{\pm 0.16}$&76.22$_{\pm 0.23}$&77.53$_{\pm 0.42}$&\underline{78.74}$_{\pm 0.07}$&78.57$_{\pm 0.24}$&78.23$_{\pm 0.34}$&77.07$_{\pm 0.38}$&77.97$_{\pm 0.14}$&\textbf{79.48}$_{\pm 0.14}$\\
			
			\hdashline
			WRN-16-2~\cite{zagoruyko2016wide}&72.55$_{\pm 0.24}$&73.12$_{\pm 0.31}$&73.51$_{\pm 0.25}$&73.99$_{\pm 0.46}$&75.46$_{\pm 0.14}$&\underline{76.07}$_{\pm 0.34}$&73.36$_{\pm 0.21}$&74.45$_{\pm 0.32}$&\textbf{77.31}$_{\pm 0.32}$ \\
			
			WRN-40-2~\cite{zagoruyko2016wide}&76.89$_{\pm 0.29}$&78.68$_{\pm 0.32}$&78.75$_{\pm 0.18}$&78.84$_{\pm 0.26}$&78.57$_{\pm 0.34}$&\underline{79.57}$_{\pm 0.39}$&78.14$_{\pm 0.35}$&78.71$_{\pm 0.25}$&\textbf{80.96}$_{\pm 0.34}$ \\

			\hdashline
			ShuffleV2 0.5$\times$~\cite{ma2018shufflenet}&67.39$_{\pm 0.35}$&69.92$_{\pm 0.15}$&\underline{71.64}$_{\pm 0.36}$&71.21$_{\pm 0.44}$&70.38$_{\pm 0.37}$&71.02$_{\pm 0.29}$&71.32$_{\pm 0.17}$&70.88$_{\pm 0.32}$&\textbf{72.61}$_{\pm 0.24}$ \\
			
			ShuffleV2 1$\times$~\cite{ma2018shufflenet}&70.93$_{\pm 0.24}$&74.09$_{\pm 0.32}$&74.93$_{\pm 0.36}$&76.55$_{\pm 0.42}$&75.85$_{\pm 0.19}$&75.31$_{\pm 0.26}$&\underline{76.73}$_{\pm 0.34}$&75.68$_{\pm 0.37}$&\textbf{77.25}$_{\pm 0.35}$ \\
			\hdashline
			HCGNet-A1~\cite{yang2020gated}&77.42$_{\pm 0.16}$&79.18$_{\pm 0.23}$&78.32$_{\pm 0.38}$&80.28$_{\pm 0.13}$&80.08$_{\pm 0.33}$&\underline{80.64}$_{\pm 0.16}$&79.91$_{\pm 0.41}$&80.27$_{\pm 0.22}$&\textbf{81.66}$_{\pm 0.14}$ \\
			
			HCGNet-A2~\cite{yang2020gated}&79.00$_{\pm 0.41}$&81.17$_{\pm 0.45}$&80.49$_{\pm 0.34}$&81.66$_{\pm 0.35
			}$&81.87$_{\pm 0.44}$&\underline{82.42}$_{\pm 0.26}$&81.83$_{\pm 0.22}$&81.62$_{\pm 0.34}$&\textbf{83.14}$_{\pm 0.25}$ \\
			\hline			
	\end{tabular}}
	
	\label{same_arch} 
\end{table*}

\begin{table*}[tbp]
	\centering
	\caption{Top-1 accuracy (\%) of online KD methods by jointly training \textbf{two networks with different architectures} on CIFAR-100. The \textbf{bold number} is the best result among various methods, while the \underline{underline number} is the second best.}
	\begin{tabular}{c|c|cc|cc|cc|cc}  
		\hline
		\multirow{2}{*}{Network}&Net1&ResNet-32&ResNet-56&WRN-16-2&WRN-40-2&ResNet-56&ResNet-110&ShuffleV2 1$\times$&ShuffleV2 1$\times$ \\ 
		&Net2&ResNet-110&ResNet-110&WRN-40-2&WRN-28-4&WRN-40-2&WRN-28-4&ResNet-110&WRN-40-2\\ 
		\hline
		\multirow{2}{*}{\#params}&Net1&0.47M&0.86M&0.70M&2.26M&0.86M&1.17M&1.36M&1.36M \\ 
		&Net2&1.17M&1.17M&2.26M&5.87M&2.26M&5.87M&1.17M&2.26M \\
		\hline
		\multirow{2}{*}{Baseline}&Net1&70.91$_{\pm 0.14}$&73.15$_{\pm 0.23}$&72.55$_{\pm 0.24}$&76.89$_{\pm 0.29}$&73.15$_{\pm 0.23}$&75.29$_{\pm 0.16}$&70.93$_{\pm 0.24}$&70.93$_{\pm 0.24}$\\ 
		&Net2&75.29$_{\pm 0.16}$&75.29$_{\pm 0.16}$&76.89$_{\pm 0.29}$&79.17$_{\pm 0.29}$&76.89$_{\pm 0.29}$&79.17$_{\pm 0.29}$&75.29$_{\pm 0.16}$&76.89$_{\pm 0.29}$\\ 
		\hline
		\multirow{2}{*}{DML~\cite{zhang2018deep}}&Net1&73.13$_{\pm 0.31}$&75.87$_{\pm 0.19}$&76.02$_{\pm 0.08}$&78.48$_{\pm 0.15}$&74.90$_{\pm 0.25}$&77.46$_{\pm 0.44}$&75.48$_{\pm 0.19}$&75.39$_{\pm 0.22}$\\ 
		&Net2&78.10$_{\pm 0.28}$&78.35$_{\pm 0.16}$&79.31$_{\pm 0.23}$&81.08$_{\pm 0.09}$&78.69$_{\pm 0.36}$&80.73$_{\pm 0.16}$&78.47$_{\pm 0.31}$&79.04$_{\pm 0.15}$\\ 
		\hline
		\multirow{2}{*}{AFD~\cite{chung2020feature}}&Net1&73.59$_{\pm 0.27}$&74.57$_{\pm 0.24}$&\underline{76.13}$_{\pm 0.36}$&78.21$_{\pm 0.22}$&75.60$_{\pm 0.19}$&77.49$_{\pm 0.34}$&75.74$_{\pm 0.22}$&75.27$_{\pm 0.35}$\\ 
		&Net2&77.99$_{\pm 0.38}$&75.48$_{\pm 0.17}$&78.96$_{\pm 0.22}$&80.42$_{\pm 0.26}$&78.52$_{\pm 0.35}$&81.16$_{\pm 0.46}$&77.20$_{\pm 0.42}$&78.65$_{\pm 0.28}$\\ 
		\hline
		\multirow{2}{*}{KDCL~\cite{guo2020online}}&Net1&73.37$_{\pm 0.17}$&76.01$_{\pm 0.42}$&75.33$_{\pm 0.37}$&78.81$_{\pm 0.26}$&74.55$_{\pm 0.36}$&78.23$_{\pm 0.26}$&75.73$_{\pm 0.17}$&75.26$_{\pm 0.31}$\\ 
		&Net2&75.63$_{\pm 0.25}$&76.49$_{\pm 0.37}$&77.19$_{\pm 0.41}$&80.51$_{\pm 0.25}$&77.07$_{\pm 0.46}$&79.62$_{\pm 0.37}$&77.45$_{\pm 0.33}$&77.27$_{\pm 0.29}$\\ 
		\hline
		\multirow{2}{*}{PCL~\cite{wu2021peer}}&Net1&\underline{75.06}$_{\pm 0.17}$&76.13$_{\pm 0.47}$&76.02$_{\pm 0.27}$&\underline{79.29}$_{\pm 0.18}$&\underline{76.32}$_{\pm 0.41}$&\underline{79.11}$_{\pm 0.36}$&\underline{76.17}$_{\pm 0.55}$&\underline{76.21}$_{\pm 0.14}$\\ 
		&Net2&\underline{78.60}$_{\pm 0.46}$&78.74$_{\pm 0.32}$&\underline{79.46}$_{\pm 0.47}$&\underline{81.63}$_{\pm 0.29}$&\underline{79.46}$_{\pm 0.27}$&\underline{82.05}$_{\pm 0.42}$&\underline{78.74}$_{\pm 0.16}$&79.32$_{\pm 0.38}$\\ 
		\hline
		\multirow{2}{*}{CKD-MKT~\cite{gou2022collaborative}}&Net1&73.35$_{\pm 0.23}$&\underline{76.21}$_{\pm 0.51}$&76.03$_{\pm 0.26}$&78.17$_{\pm 0.42}$&75.31$_{\pm 0.36}$&77.76$_{\pm 0.52}$&75.37$_{\pm 0.22}$&75.22$_{\pm 0.24}$\\ 
		&Net2&78.48$_{\pm 0.33}$&\underline{78.77}$_{\pm 0.12}$&79.17$_{\pm 0.16}$&80.40$_{\pm 0.23}$&78.85$_{\pm 0.47}$&81.47$_{\pm 0.28}$&78.39$_{\pm 0.27}$&\underline{79.55}$_{\pm 0.36}$\\ 
		\hline
		\multirow{2}{*}{L-MCL (Ours)}&Net1&\textbf{75.82}$_{\pm 0.24}$&\textbf{77.04}$_{\pm 0.21}$&\textbf{76.88}$_{\pm 0.31}$&\textbf{80.42}$_{\pm 0.14}$&\textbf{77.54}$_{\pm 0.26}$&\textbf{80.02}$_{\pm 0.26}$&\textbf{76.85}$_{\pm 0.14}$&\textbf{77.06}$_{\pm 0.41}$\\ 
		&Net2&\textbf{79.71}$_{\pm 0.16}$&\textbf{79.56}$_{\pm 0.26}$&\textbf{80.24}$_{\pm 0.22}$&\textbf{82.61}$_{\pm 0.34}$&\textbf{80.72}$_{\pm 0.17}$&\textbf{82.84}$_{\pm 0.29}$&\textbf{79.17}$_{\pm 0.35}$&\textbf{80.29}$_{\pm 0.22}$\\  
		\hline
	\end{tabular}
	
	\label{cross_different} 
\end{table*}

\subsection{Results on CIFAR-100 Classification}
\textbf{CIFAR-100 Dataset.} CIFAR-100~\cite{krizhevsky2009learning} is a classification dataset composed of natural images. It includes 50K training images and 10K test images drawn from 100 classes. As the common protocol, we follow the standard data augmentation and preprocessing pipeline~\cite{huang2017densely}, \emph{i.e.} random cropping and flipping. The input size of each image is 32$\times$32.

\textbf{Implementation details.} All networks are
trained by SGD with a momentum of 0.9, a batch size of 128 and a weight decay of $5\times 10^{-4}$. We use a cosine learning rate that starts from 0.1 and gradually decreases to 0 throughout the 300 epochs. As suggested by Chen \emph{et al.}~\cite{chen2020simple}, we use $\tau=0.1$ for similarity calibration and $d=128$ as the contrastive embedding size. We adopt mini-batch-based contrastive sample mining and set $K=126$ (\emph{i.e.} batch size$-2$) as the number of negative samples. We report the average result with the standard deviation (average$\pm$std) over three runs for a fair comparison. \emph{More analyses about hyper-parameter strategies are shown in Section~\ref{Ablation}.}

\textbf{Comparison with SOTA online KD methods under two peer networks with the same architecture}. As shown in Table~\ref{same_arch}, we first investigate the effectiveness of L-MCL, an improved method of our previous proposed MCL~\cite{yang2022mutual}. Some popular backbone networks for image classification, such as ResNets~\cite{he2016deep},
WRNs~\cite{zagoruyko2016wide}, HCGNets~\cite{yang2020gated} and ShuffleNetV2~\cite{ma2018shufflenet}, are utilized to evaluate the performance. Moreover, many representative online KD methods are compared, such as DML~\cite{zhang2018deep}, ONE~\cite{zhu2018knowledge}, AFD~\cite{chung2020feature}, KDCL~\cite{guo2020online} and PCL~\cite{wu2021peer}, to verify the superiority of our L-MCL. All shown results in Table~\ref{same_arch} are achieved from jointly training two networks with the same architecture. As expected, all online KD methods improve the classification performance consistently across various network architectures compared to the independent training method. The results indicate that a network can benefit from peer-teaching with another network. 

Our proposed L-MCL outperforms the previous SOTA PCL with average accuracy gains of 1.28\% and 1.31\% and 1.77\% and  0.86\% on ResNets, WRNs, ShuffleNet and HCGNet families, respectively. Moreover, it is hard to say which is the \emph{second-best} method since different methods are superior for various architectures or datasets. Previous online KD approaches can be concluded to focus on distilling logit-level class probability among multiple networks but mainly differ in various learning strategies. However, exploring a single type of logit-level knowledge may limit the performance improvement for online KD methods. Beyond logit-level distillation, our L-MCL aims at contrastive representation learning by taking advantage of collaborative learning. Compared to the baseline, L-MCL achieves average accuracy improvements of 4.49\%, 4.41\%, 5.77\% and 4.19\% on ResNets, WRNs, ShuffleNet and HCGNet families, respectively. The results suggest that L-MCL can help each network in the cohort learn better representations, conducive to classification performance.

\begin{table*}[tbp]
	\centering
	\caption{Top-1 accuracy (\%) of jointly training \textbf{three networks with the same architecture} on CIFAR-100. The \textbf{bold number} represents the best result among various methods, while the \underline{underline number} denotes the second best.}
	\begin{tabular}{c|ccccccc|c}  
		\hline
		Network&Baseline&DML~\cite{zhang2018deep}&ONE~\cite{zhu2018knowledge}&OKDDip~\cite{chen2020online}&AFD~\cite{chung2020feature}&KDCL~\cite{guo2020online}&PCL~\cite{wu2021peer}&L-MCL (Ours) \\ 
		\hline
		ResNet-32~\cite{he2016deep}&70.91$_{\pm 0.14}$&74.48$_{\pm 0.33}$&73.82$_{\pm 0.24}$&74.19$_{\pm 0.17}$&72.91$_{\pm 0.34}$&73.84$_{\pm 0.26}$&\underline{74.67}$_{\pm 0.34}$&\textbf{76.35}$_{\pm 0.27}$\\
		
		ResNet-56~\cite{he2016deep}&73.15$_{\pm 0.23}$&75.71$_{\pm 0.18}$&75.77$_{\pm 0.28}$&76.28$_{\pm 0.24}$&76.11$_{\pm 0.16}$&75.38$_{\pm 0.42}$&\underline{76.68}$_{\pm 0.17}$&\textbf{77.72}$_{\pm 0.32}$\\
		
		\hdashline
		WRN-16-2~\cite{zagoruyko2016wide}&72.55$_{\pm 0.24}$&\underline{76.13}$_{\pm 0.28}$&74.04$_{\pm 0.41}$&73.80$_{\pm 0.21}$&74.08$_{\pm 0.38}$&75.97$_{\pm 0.19}$&75.55$_{\pm 0.34}$&\textbf{77.48}$_{\pm 0.25}$ \\
		
		WRN-40-2~\cite{zagoruyko2016wide}&76.89$_{\pm 0.29}$&78.63$_{\pm 0.32}$&79.12$_{\pm 0.37}$&79.18$_{\pm 0.33}$&79.07$_{\pm 0.16}$&78.42$_{\pm 0.39}$&\underline{80.00}$_{\pm 0.25}$&\textbf{81.21}$_{\pm 0.35}$ \\
		
		\hdashline
		
		ShuffleNetV2 0.5$\times$~\cite{ma2018shufflenet}&67.39$_{\pm 0.35}$&71.72$_{\pm 0.27}$&71.29$_{\pm 0.38}$&71.14$_{\pm 0.42}$&\underline{71.96}$_{\pm 0.25}$&71.43$_{\pm 0.16}$&71.74$_{\pm 0.27}$ &\textbf{73.21}$_{\pm 0.16}$ \\
		
		ShuffleNetV2 1$\times$~\cite{ma2018shufflenet}&70.93$_{\pm 0.24}$&75.99$_{\pm 0.34}$&75.52$_{\pm 0.29}$&75.61$_{\pm 0.23}$&\underline{76.95}$_{\pm 0.26}$&75.71$_{\pm 0.25}$&76.36$_{\pm 0.14}$ &\textbf{77.56}$_{\pm 0.25}$ \\
		\hline
	\end{tabular}
	\label{three_nets} 
\end{table*}

\begin{figure*}[tbp]  
	\centering
	\includegraphics[width=1\linewidth]{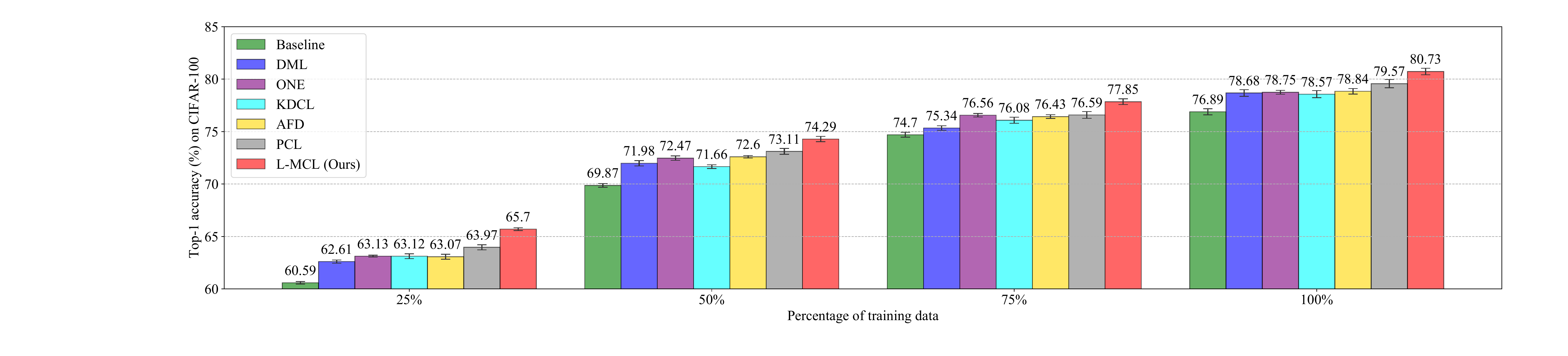}
	\caption{Top-1 accuracy (\%) of various online KD methods under few-shot scenario with different percentages of training data. The performance is evaluated on the WRN-40-2 backbone. We retain 25\%, 50\%, 75\% and 100\% samples of the training set, respectively. We maintain the original test set unchanged.}
	\label{sp_few_shot}
\end{figure*}

\textbf{Comparison with SOTA online KD methods under two networks with different architectures}. As shown in Table~\ref{cross_different}, we further conduct experiments on two different network architectures for online KD methods, where 'Net2' represents a higher-capacity network than 'Net1'. We observe some similar conclusions on performance improvements compared with the scenario of the same network pair. All online KD methods generally improve the performance across various combinations of network pairs. The results reveal that existing online KD methods often do not rely on architecture-specific cues. Our L-MCL achieves the best accuracy gains against other approaches. It also surpasses the best competitor PCL over eight network pairs with average accuracy improvements of 0.92\% and 0.89\% on Net1 and Net2, respectively. The results verify the scalability of our L-MCL to work well on the different-architecture setup.

Besides, we also find other observations about L-MCL to deal with two different networks. A higher-capacity network (Net2) can still obtain a significant performance improvement even combined with a lower network (Net1) as the partner. For example, ResNet-110 gets a 4.42\% gain when combined with a ResNet-32. Notice that the baseline performance gap between ResNet-32 (70.91\%) and ResNet-110 (75.29\%) is quite large. A similar observation also occurs in the WRN family. WRN-40-2 obtains a 3.35\% improvement when a WRN-16-2 network is selected as the partner. Here, the performance comparison of WRN-16-2 v.s WRN-40-2 is 72.55\% v.s 76.89\%. Moreover, we observe that the performance improvement of a given network may be approximate by partnering with various networks of different capacities. For example, ResNet-110 shows 4.42\%, 4.27\%, 4.73\% and 3.88\% accuracy gains when combined with ResNet-32, ResNet-56, WRN-28-4 and ShuffleNetV2 respectively. Analogously, WRN-40-2 achieves 3.35\%, 3.53\%, 3.83\% and 3.40\% accuracy improvements when WRN-16-2, WRN-28-4, ResNet-56 and ShuffleNetV2 are used as partners, respectively. This superiority allows us to apply L-MCL for resource-constrained scenarios to enhance the target network by selecting a lightweight network as the partner.

\textbf{Extend online KD methods to three networks}. As shown in Table~\ref{three_nets}, we experiment with online KD methods on three networks. As the number of networks in the cohort increases, most approaches generally lead to better accuracy gains than the scenario of two networks. This is because more networks may capture richer knowledge for collaborative learning. Our L-MCL consistently achieves the best performance and outperforms the previous SOTA PCL with 1.36\%, 1.52\% and 1.34\% average gains on ResNet, WRN and ShuffleNet families, respectively. The results further verify our claim that performing contrastive representation learning is an effective way for online KD. It seems that the performance gains of online KD methods may saturate at three networks. For example, L-MCL applied to three networks only achieves an average increase of 0.35\% across six network architectures compared to the counterpart of two networks.

\textbf{Comparsion under few-shot scenario.} In practice, available training samples may be scarce. As shown in Fig.~\ref{sp_few_shot}, we conduct experiments on various online KD methods under few-shot scenarios by maintaining 25\%, 50\% and 75\% training samples. We adopt stratified sampling to make the newly crafted training set class-balanced. For a fair comparison, we utilize the same data split strategy for each few-shot ratio while reserving the original test set unchanged. Our L-MCL can outperform other competitive approaches consistently across various few-shot ratios. It surpasses the best competitor PCL with 1.73\%, 1.18\% and 1.26\% margins when trained with 25\%, 50\% and 75\% training samples, respectively. The results verify that L-MCL can generalize better to the few-shot scenario, which may be attributed to the contrastive learning for learning general features. In contrast, previous online KD approaches often rely on class probability distillation, which may overfit the limited set yet generalize worse to the test set.

\begin{table}[tbp]
	\centering
	\caption{Top-1 accuracy (\%) on STL-10 and TinyImageNet under the linear classification protocol. We freeze the feature extractor pre-trained on CIFAR-100 and train a linear classifier over features after global average pooling.}
	\resizebox{1.\linewidth}{!}{
		\begin{tabular}{l|cc|cc}  
			\hline 
			\multirow{2}{*}{Method}&\multicolumn{2}{c|}{CIFAR-100$\rightarrow$STL-10} &\multicolumn{2}{c}{CIFAR-100$\rightarrow$TinyImageNet} \\  
			&ResNet-56 & WRN-40-2&ResNet-56 & WRN-40-2 \\  
			\hline
			Baseline&60.54&62.11&22.53&23.24 \\
			DML~\cite{zhang2018deep}&58.78&61.02&22.33&22.40 \\
			ONE~\cite{zhu2018knowledge}&61.45&64.18&23.37&24.36 \\
			KDCL~\cite{guo2020online}&60.30&59.95&22.93&23.24 \\
			AFD~\cite{chung2020feature}&60.54&59.91&22.96&21.95 \\
			PCL~\cite{wu2021peer}&62.81&62.16&23.64&23.85 \\
			\hline
			MCL~\cite{yang2022mutual}&63.13&64.75&24.53&25.32 \\
			L-MCL&\textbf{64.52}&\textbf{67.86}&\textbf{26.82}&\textbf{28.74}\\
			\hline
	\end{tabular}}
	
	\label{transfer_img_cls}
\end{table}

\subsection{Results of Transfer Learning to Image Classification on STL-10 and TinyImageNet}
\textbf{STL-10 and TinyImageNet Dataset.} STL-10~\cite{CoatesNL11} contains 5K
labeled training images and 8K test images in 10 classes.
TinyImageNet\footnote{http://tiny-imagenet.herokuapp.com/} includes 100K training images and
10k test images in 200 classes. Each input image is downsampled to 32$\times$32 to match the size of CIFAR-100 under the transfer learning setup. And we follow the data augmentation and preprocessing pipeline as same as CIFAR-100.

\textbf{Implementation details.} We freeze the feature extractor pre-trained on CIFAR-100 and train a linear classifier over the transferred dataset. We utilize a SGD
optimizer with a momentum of 0.9, a batch size of 64 and a weight decay of 0. The initial learning rate starts at 0.1 and is decayed by 10 at the 30-th, 60-th and 90-th epochs within the total 100 epochs.

\textbf{Comparsion of transfer learning to STL-10 and TinyImageNet.} A desirable property of online KD methods is to guide the network to acquire \emph{general feature representations} that can well transfer to other unseen semantic recognition tasks. We follow the linear classification protocol~\cite{he2020momentum} to quantify the transferability of features. A network is pre-trained by CIFAR-100 and served as a frozen feature extractor. We train a supervised linear classifier (a fully-connected layer followed by softmax) to perform 10-way (for STL-10) or 200-way (for TinyImageNet) classification. 

As shown in Table~\ref{transfer_img_cls}, we observe that the current online KD methods cannot improve the downstream classification performance effectively compared to the baseline. We conjecture that logit-based distillation might make the network biased  towards the original task. In contrast, our MCL or L-MCL can significantly enhance the downstream classification. L-MCL achieves the best results and outperforms the baseline with average margins of 4.86\% and 4.89\% on STL-10 and TinyImageNet, respectively. The results reveal that our L-MCL can lead the network to learn well-generalized feature representations.

\begin{table}[tbp]
	\centering
	\caption{Top-1 accuracy (\%) of online KD methods by jointly training \textbf{two networks with the same architecture} for ImageNet classification. The \textbf{bold number} represents the best result among various methods, while the \underline{underline number} denotes the second best.}
	\resizebox{1.\linewidth}{!}{
		\begin{tabular}{l|ccccc}  
			\hline 
			Network&ResNet-18&ResNet-34 &ResNet-50&ShuffleV2 \\  
			\#Params&11.69M&21.80M&25.56M&2.28M \\  
			\hline
			Baseline&69.95&73.68&76.28&64.25 \\
			DML~\cite{zhang2018deep}&70.60&74.59&77.21&65.12 \\
			ONE~\cite{zhu2018knowledge}&70.56&74.29 &76.95&\underline{65.35} \\
			KDCL~\cite{guo2020online}&70.13&74.31&77.06&64.14 \\
			AFD~\cite{chung2020feature}&69.96&74.50&76.84&64.47  \\
			PCL~\cite{wu2021peer}&70.64&74.29&76.77&64.45 \\
			FFSD~\cite{li2022distilling}&\underline{70.89}&\underline{74.66}&77.08&64.85\\
			CKD-MKT~\cite{gou2022collaborative}&70.78&74.55&\underline{77.31}&65.02 \\
			\hline
			MCL~\cite{yang2022mutual}&70.67&74.51&77.24&65.14 \\
			L-MCL(Ours)&\textbf{71.38}&\textbf{75.12 }&\textbf{78.35}&\textbf{65.91}\\
			\hline
	\end{tabular}}
	\label{imagenet_same_net}
\end{table}

\begin{table}[tbp]
	\centering
	\caption{Top-1 accuracy (\%) of online KD methods by jointly training \textbf{two networks with different architectures} for ImageNet classification. The \textbf{bold number} represents the best result among various methods, while the \underline{underline number} denotes the second best.}
	\begin{tabular}{l|c|ccc}  
		\hline 
		\multirow{2}{*}{Network}&Net1&ShuffleV2&ShuffleV2 &ResNet-18\\  
		&Net2&ResNet-18&ResNet-50 &ResNet-50\\  
		\hline
		\multirow{2}{*}{\#Params}&Net1&2.28M&2.28M&11.69M \\  
		&Net2&11.69M&25.56M&25.56M \\  
		\hline
		\multirow{2}{*}{Baseline}&Net1&64.25&64.25&69.95\\
		&Net2&69.95&76.28&76.28\\
		\hline
		\multirow{2}{*}{DML~\cite{zhang2018deep}}&Net1&\underline{65.35}&65.34&\underline{71.03} \\
		&Net2&70.13&75.46&76.27 \\
		\hline
		\multirow{2}{*}{KDCL~\cite{guo2020online}}&Net1&64.58&64.49&70.34 \\
		&Net2&\underline{70.43}&75.41&\underline{76.58} \\
		\hline
		\multirow{2}{*}{AFD~\cite{chung2020feature}}&Net1&64.72&65.42&70.85\\
		&Net2&69.98&75.07&75.34 \\
		\hline
		\multirow{2}{*}{PCL~\cite{wu2021peer}}&Net1&65.29&63.59&70.08 \\
		&Net2&70.24&\underline{76.20}&75.93  \\
		\hline
		\multirow{2}{*}{CKD-MKT~\cite{gou2022collaborative}}&Net1&64.52&\underline{65.43}&71.02 \\
		&Net2&70.34&75.80&76.52\\
		\hline
		\multirow{2}{*}{L-MCL(Ours)}&Net1&\textbf{66.06}&\textbf{66.44}&\textbf{71.69}\\
		&Net2&\textbf{71.16}&\textbf{76.57 }&\textbf{77.34}\\
		\hline
	\end{tabular}
	\label{imagenet_diff_net}
\end{table}

\begin{table}[tbp]
	\centering
	\caption{Top-1 accuracy (\%) of online KD methods by jointly training \textbf{two networks with the same architecture} on Swin Transformer~\cite{liu2021swin} for ImageNet classification.}
	\begin{tabular}{l|cc}  
		\hline 
		Network&Swin-Tiny&Swin-Small \\  
		\#Params&11.69M&21.80M \\  
		\hline
		Baseline&81.28&82.94\\
		DML~\cite{zhang2018deep}&81.46&83.12 \\
		\hline
		MCL~\cite{yang2022mutual}&81.74&83.26 \\
		L-MCL(Ours)&\textbf{82.14}&\textbf{83.48}\\
		\hline
	\end{tabular}
	\label{vit_same_net}
\end{table}

\subsection{Results on ImageNet Classification}
\textbf{ImageNet.} ImageNet~\cite{deng2009imagenet} is a large-scale image classification dataset that contains 1.2 million training images and 50K validation images in 1000 classes. As the common protocol, we follow the standard data augmentation and preprocessing pipeline~\cite{huang2017densely}. The input size of each image is 224$\times$224.

\textbf{Implementation details.}
All CNN networks are trained by SGD with a momentum of 0.9, a batch size of 256 and a weight decay of $1\times 10^{-4}$. The initial learning rate starts at 0.1 and is decayed by 10 at 30 and 60 epochs within the total 90 epochs. As suggested by Wu \emph{et al.}~\cite{wu2018unsupervised}, we use  $\tau=0.07$ on ImageNet for similarity calibration and $d=128$ as the contrastive embedding size. We adopt memory-based contrastive sample mining, retrieving one positive and $K=8192$ negative embeddings from the online memory bank. The training setup of Swin Transformer are followed by the original paper~\cite{liu2021swin}. \emph{More analyses about hyper-parameter strategies are shown in Section~\ref{Ablation}.}

\textbf{Comparison with SOTA online KD methods under two peer networks with the same architecture}. As shown in Table~\ref{Detection}, we further conduct experiments on the more challenging ImageNet to compare various online KD approaches applied to serveral networks, including ResNet family~\cite{he2016deep} and light-weight ShuffleNetV2~\cite{ma2018shufflenet}. Our L-MCL shows the best classification performance on ImageNet among OKD methods. Compared with the baseline, L-MCL achieves 1.43\%, 1.44\%, 2.07\% and 1.66\% accuracy gains on ResNet-18, ResNet-34, ResNet-50 and ShuffleV2, respectively. It also beats the SOTA CKD-MKT~\cite{gou2022collaborative} with an average improvement of 0.78\%. The result demonstrates the scalability of mutual contrastive learning for OKD on the large-scale dataset. L-MCL outperforms the previous MCL with an average gain of 0.80\%, demonstrating the superiority of layer association for MCL.

\textbf{Comparison with SOTA online KD methods under two networks with different architectures}. As shown in Table~\ref{imagenet_diff_net}, we further conduct experiments on two different network architectures for online KD methods, where 'Net2' represents a higher-capacity network than 'Net1'. Our L-MCL achieves the best classification performance among online KD methods and results in consistent gains over baselines across different network pairs. When trained by L-MCL combined with ResNet-50, ShuffleNetV2 and ResNet-18 achieve the best 66.44\% and 71.69\% accuracy performance, outperforming baseline with 2.19\% and 1.74\%, and SOTA CKD-MKT with 1.01\% and 0.67\%. We also find that Net1 may lead to a negative impact for Net2 when the capacity gap between them is large, existing in previous methods. For example, ResNet-50 underperforms baseline with 0.08\%$\sim$1.21\% points when combined with the small ShuffleNetV2 trained by previous online KD approaches. In contrast, our L-MCL robustly boosts network pairs for joint training. The results verify the scalability of our L-MCL to work well on the different-architecture setup for the large-scale ImageNet.

\textbf{Apply L-MCL to Vision Transformer}. Beyond CNN, we also extend L-MCL to Swin Transformer~\cite{liu2021swin}, a representative powerful vision transformer architecture. We also compare logit-level DML~\cite{zhang2018deep} method. Notice that other online KD methods are designed for CNN and cannot be extended to vision transformer. As shown in Table~\ref{vit_same_net}, L-MCL can achieve 0.86\% and 0.54\% improvements on Swin-Tiny and Swin-Small, even if the baseline performance is already high. In contrast, the logit-level DML only leads to marginal gains around 0.1\%$\sim$0.2\% points. The results reveal that mutual contrastive learning is also critical for vision-transformer-based online KD.

\begin{table}[tbp]
	\centering
	\caption{MAP (\%) of online KD methods on transfer learning to COCO-2017 based on Mask-RCNN framework for object detection and instance segmentation. The pretrained ResNet backbones are borrowed from Table~\ref{imagenet_same_net}. The \textbf{bold number} represents the best result among various methods, while the \underline{underline number} denotes the second best.}
	\begin{tabular}{l|cc|cc|cc}  
		\hline 
		\multirow{2}{*}{Method} &\multicolumn{2}{c|}{ResNet-18}&\multicolumn{2}{c|}{ResNet-34}&\multicolumn{2}{c}{ResNet-50} \\  
		&bbox & segm	&bbox & segm	&bbox & segm \\  
		\hline
		Baseline&33.4&30.2&35.1&31.8& 36.9&33.4\\
		DML~\cite{zhang2018deep}&33.7&30.5&36.4&32.8&37.7&34.2 \\
		ONE~\cite{zhu2018knowledge}&33.8&30.7&35.7&32.3 &37.3&33.8\\
		KDCL~\cite{guo2020online}&33.2&30.1&35.5&32.1 &37.2&33.5\\
		AFD~\cite{chung2020feature}&33.1&29.8&35.9&32.5&36.9&33.3 \\
		PCL~\cite{wu2021peer}&33.9&31.1&35.6&32.4&37.0&33.5 \\
		FFSD~\cite{li2022distilling}&34.1&31.4&\underline{36.7}&\underline{33.1}&37.3&33.8 \\
		CKD-MKT~\cite{gou2022collaborative}&33.8&30.9&36.5&32.7&37.9&34.4 \\
		\hline
		MCL~\cite{yang2022mutual}&\underline{34.2}&\underline{31.3}&36.2&32.5&\underline{38.0}&\underline{34.6} \\
		L-MCL(Ours)&\textbf{35.0}&\textbf{32.1}&\textbf{37.2}&\textbf{33.7}&\textbf{38.5}&\textbf{35.0}\\
		\hline
	\end{tabular}
	\label{Detection}
\end{table}

\subsection{Results of Transfer Learning to Object Detection and Instance Segmentation on COCO-2017}
\textbf{COCO-2017 Dataset.} COCO-2017~\cite{lin2014microsoft} is a widely adopted dataset for object detection and instance segmentation. It includes 120k training images and 5k validation images with 80 object categories.

\textbf{Implementation details.} We conduct object detection experiments based on the MMDetection~\cite{chen2019mmdetection}. All experiments are implemented on 8 GPUs using synchronized SGD with two images per GPU. We adopt a 1x learning scheduler with 12 epochs. Other setups are followed by the default MMDetection.

\textbf{Comparison of transfer learning to object detection and instance segmentation on COCO-2017.} We take pre-trained ResNets in Table~\ref{imagenet_same_net} on ImageNet using various online KD methods as the backbone over Mask-RCNN~\cite{he2017mask} for downstream object detection and instance segmentation. As shown in Table~\ref{Detection}, using L-MCL to train the ResNet feature extractors on ImageNet achieves significant mAP gains with average 1.7\% and 1.8\% points for downstream detection and segmentation tasks compared to the baseline. Moreover, our L-MCL also outperforms SOTA CKD-MKT~\cite{gou2022collaborative} with average 0.8\% and 0.9\% mAP improvements on object detection and instance segmentation, respectively. The results demonstrate the efficacy of L-MCL for learning better representations for downstream semantic recognition tasks.

\subsection{Ablation Study and Parameter Analysis}
\label{Ablation}

\begin{table}[tbp]
	\centering
	\caption{Ablation study of loss terms in L-MCL over the setup of the same network pairs on CIFAR-100.}
	\resizebox{1.\linewidth}{!}{
		\begin{tabular}{cccc|cc}  
			\hline 
			$\mathcal{L}^{VCL}$&$\mathcal{L}^{Soft\_VCL}$&$\mathcal{L}^{ICL}$&$\mathcal{L}^{Soft\_ICL}$&ResNet-110&WRN-40-2 \\  
			\hline
			-&-&-&-&75.29$_{\pm 0.16}$&76.89$_{\pm 0.29}$ \\
			\checkmark&-&-&-&77.35$_{\pm 0.32}$&77.66$_{\pm 0.18}$ \\
			\checkmark&\checkmark&-&-&77.63$_{\pm 0.27}$&78.35$_{\pm 0.23}$ \\
			-&-&\checkmark&-&78.34$_{\pm 0.18}$&79.29$_{\pm 0.36}$\\
			-&-&\checkmark&\checkmark&78.83$_{\pm 0.17}$&79.56$_{\pm 0.27}$\\
			\checkmark&\checkmark&\checkmark&\checkmark&\textbf{79.05}$_{\pm 0.24}$&\textbf{79.84}$_{\pm 0.33}$\\
			\hline
	\end{tabular}}
	\label{ablation_l_mcl_loss}
\end{table}

\begin{table}[tbp]
	\centering
	\caption{Ablation study of L-MCL and logit-based Online KD (OKD) losses over the setup of the same network pairs on CIFAR-100.}
	\resizebox{1.\linewidth}{!}{
		\begin{tabular}{c|c|c|c|cc}  
			\hline 
			Baseline&L-MCL &\multicolumn{2}{c|}{Logit-level OKD}&\multirow{2}{*}{ResNet-110}&\multirow{2}{*}{WRN-40-2} \\ 
			$\mathcal{L}^{task}$&$\mathcal{L}^{L\_MCL}$&$\mathcal{L}^{task\_G}$&$\mathcal{L}^{logit}$ \\  
			\hline
			\checkmark&-&-&-&75.29$_{\pm 0.16}$&76.89$_{\pm 0.29}$ \\
			\checkmark&\checkmark&-&-&79.05$_{\pm 0.24}$&79.84$_{\pm 0.33}$\\
			\checkmark&-&-&\checkmark&76.46$_{\pm 0.28}$&77.85$_{\pm 0.23}$ \\
			\checkmark&-&\checkmark&\checkmark&77.13$_{\pm 0.31}$&78.64$_{\pm 0.24}$ \\
			\checkmark&\checkmark&\checkmark&\checkmark&\textbf{79.48}$_{\pm 0.14}$&\textbf{80.96}$_{\pm 0.34}$\\
			\hline
	\end{tabular}}
	\label{ablation_loss}
\end{table}

\begin{table}[t]
	\centering
	\caption{Ablation study of various layer-matching mechanisms over the heterogeneous network pair of ShuffleV2-ResNet-110 on CIFAR-100.}
	\begin{tabular}{c|cc}  
		\hline 
		Layer-matching mechanism&ShuffleV2&ResNet-110 \\  
		\hline
		one-to-one &74.68$_{\pm 0.26}$&77.12$_{\pm 0.25}$ \\
		all-to-all &75.29$_{\pm 0.31}$&77.62$_{\pm 0.19}$ \\
		weighted all-to-all &\textbf{76.46}$_{\pm 0.12}$&\textbf{78.85}$_{\pm 0.33}$ \\
		
		\hline
	\end{tabular}
	\label{ablation_l_mcl_embedding}
\end{table}

\textbf{Ablation study of loss terms in L-MCL.} As shown in Table~\ref{ablation_l_mcl_loss}, we observe that each loss term about contrastive learning is conducive to the performance gain. Compared to the baseline, ICL achieves 3.05\% and 2.40\% accuracy improvements on ResNet-110 and WRN-40-2, while the conventional VCL obtains 2.06\% and 0.77\%. This is because ICL is more informative than VCL by aggregating cross-network embeddings and aims to maximize the mutual information among networks. We further compare VCL and ICL coupled with their derived soft labels. The loss combination of $\mathcal{L}^{ICL}+\mathcal{L}^{Soft\_ICL}$ outperforms the counterpart of $\mathcal{L}^{VCL}+\mathcal{L}^{Soft\_VCL}$ with 1.20\% and 1.21\% accuracy margins on ResNet-110 and WRN-40-2, respectively. The results verify our claim that ICL and its soft labels are more crucial than the conventional VCL and its soft labels.  Finally, summarizing VCL and ICL into a unified framework can maximize the performance gain for collaborative representation learning.

\textbf{Ablation study of loss terms of feature-based L-MCL and logit-based online KD.} Beyond feature-level L-MCL, we also conduct logit-level online KD for joint learning. As shown in Table~\ref{ablation_loss}, we examine the contribution of each component in online KD. Compared to the baseline, applying L-MCL (\emph{i.e.} $\mathcal{L}^{L\_MCL}$) leads to 3.76\% and 2.95\% accuracy gains on ResNet-110 and WRN-40-2, while our proposed logit-level online KD (\emph{i.e.} $\mathcal{L}^{task\_G}+\mathcal{L}^{logit}$) achieves 1.84\% and 1.75\%.
The result reveals that feature-based L-MCL contributes more improvements than logit-based online KD. Without the gated mechanism (\emph{i.e.} w/o $\mathcal{L}^{task\_G}$), the logit-level online KD degenerates into a simple average aggregation of ensemble logits (\emph{i.e.} $w_{m}^{[l]}=1/L$ in Eq.~(\ref{w_ens})). It drops 0.67\% and 0.79\% margins on ResNet-110 and WRN-40-2, respectively. The results demonstrate the superiority of weighted aggregation   using a gated module. Finally, combining feature-based L-MCL and logit-based online KD into a unified framework can maximize the performance gains to 4.19\% and 4.07\%.

\textbf{Ablation study of various layer-matching mechanisms.} A crucial distinction of L-MCL over the previous MCL is various feature embeddings from different stages are used for contrastive learning. Feature embeddings generated from different convolutional stages often encode various patterns of representational information~\cite{yang2021hierarchical}. This motivates us to perform layer-wise MCL of intermediate features beyond the final embeddings. We investigate three manners of layer-matching between two networks $f_{a}$ and $f_{b}$:

\begin{enumerate}
	\item one-to-one: MCL between the same-staged intermediate layers, \emph{i.e.},
	\begin{equation}
		\lambda_{a,b}^{la,lb}=\left\{\begin{matrix}
		1,la=lb\\
		0,la\ne lb
		\end{matrix}\right.
	\end{equation}.
	\item all-to-all: MCL among all layer pairs, , \emph{i.e.}, $\lambda_{a,b}^{la,lb}=1$.
	\item weighted-all-to-all: MCL among all layer pairs with the weight $\lambda_{a,b}^{la,lb}$ learned by meta-optimization.
\end{enumerate}
Notice that $\lambda_{a,b}^{la,lb} $ is the layer-matching weight defined in Eq.(\ref{L_MCL}).

 As shown in Table~\ref{ablation_l_mcl_embedding}, weighted-all-to-all achieves the best performance among three layer-matching mechanisms, and outperforms others with significant margins.  Theoretically, the mechanisms of (1) one-to-one and (2) all-to-all are specific cases of weighted-all-to-all from the perspective of $\lambda$ optimization. From the view of feature abstraction, matched layers between two different networks may have semantic gaps for mutual contrastive learning. Therefore, performing the weighted-all-to-all mechanism to finish adaptive layer association is more reasonable.
 
\begin{figure}[t]
	\centering 
	
	\begin{subfigure}[t]{0.4\textwidth}
		\centering
		\includegraphics[width=\textwidth]{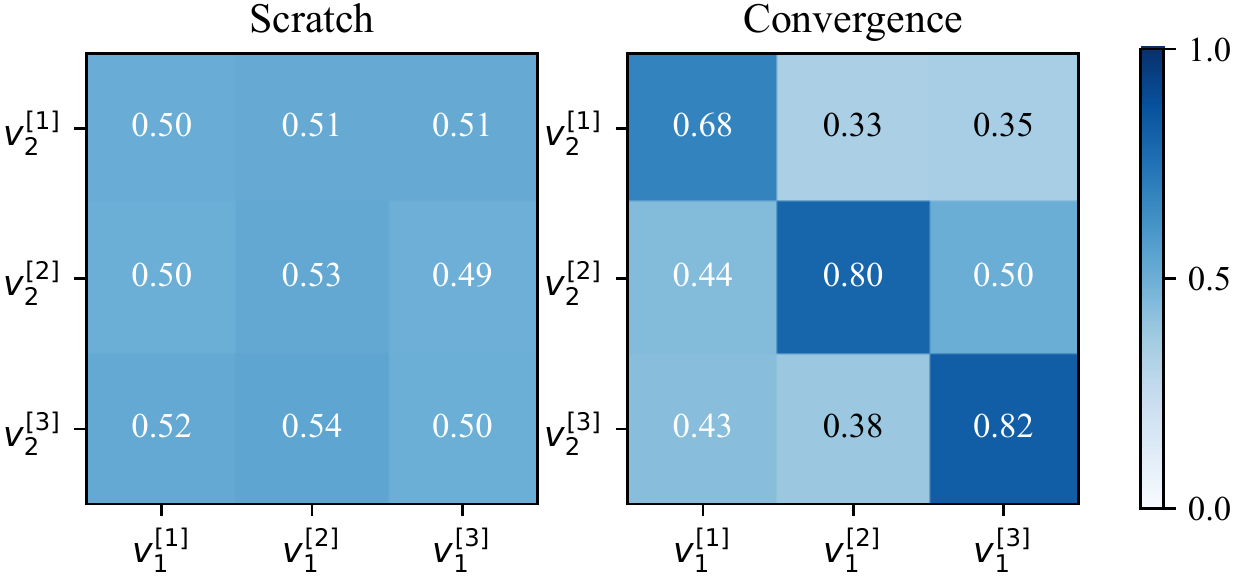}
		\caption{$\lambda$ distribution on ResNet32-ResNet110.}
		\label{res32_res110}
	\end{subfigure}
	\begin{subfigure}[t]{0.4\textwidth}
		\centering
		\includegraphics[width=\textwidth]{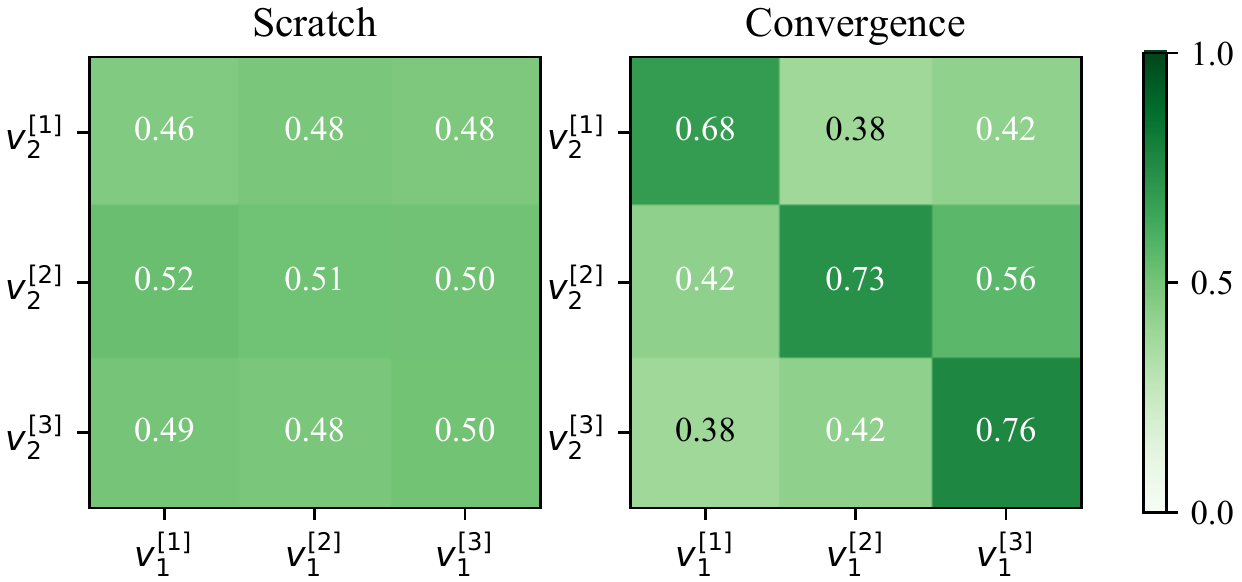}
		\caption{$\lambda$ distribution on ResNet110-ShuffleNetV2.}
		\label{res110_shuffle}
	\end{subfigure}
	\caption{Statistics of layer-matching weights $\lambda$ on the same or different architecture pairs. The weight is computed by averaging all training samples.}
	\label{lambda}
\end{figure}

\textbf{Analysis of learned layer-matching weights.} To further prove the effectiveness of the proposed layer-matching mechanism, we show the statistics of matching weights $\lambda$ on the same (ResNet32-ResNet110) or different (ResNet110-ShuffleNetV2) architecture pairs in Fig.~\ref{lambda}. We find that the mechanism tends to match the same-level layers with large weights while also assigning moderate weights to other layers. This reveals that our weighted mechanism with meta-optimization adaptively captures the semantic relations between paired feature layers in a data-driven way, compared to the manual layer-matching. This advantage allows our method readily to extend to any network pair without consideration of architectural differences.

\begin{figure}[t]
	\centering 
	\begin{subfigure}[t]{0.47\textwidth}
		\centering
		\includegraphics[width=\textwidth]{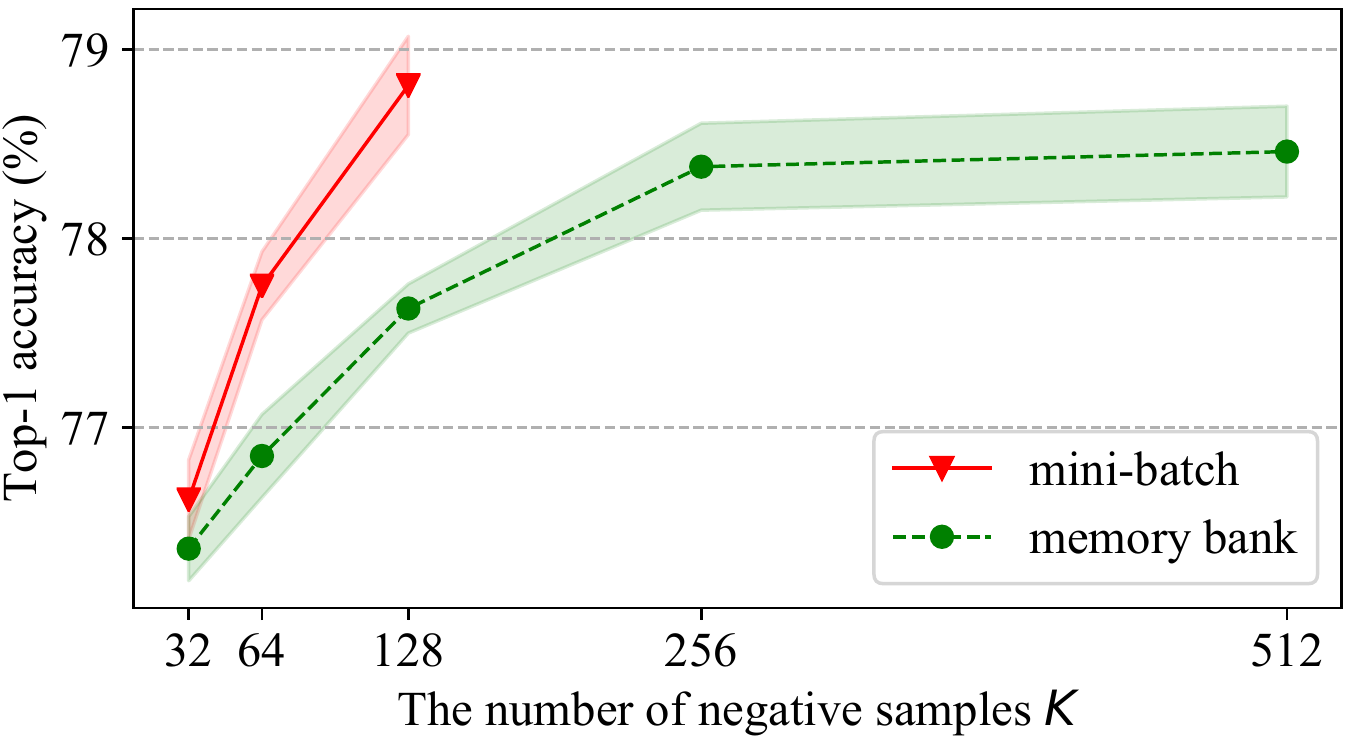}
		\caption{Top-1 accuracy (\%) over ResNet-110 on CIFAR-100.}
		\label{cifar_neg}
	\end{subfigure}
	\begin{subfigure}[t]{0.47\textwidth}
		\centering
		\includegraphics[width=\textwidth]{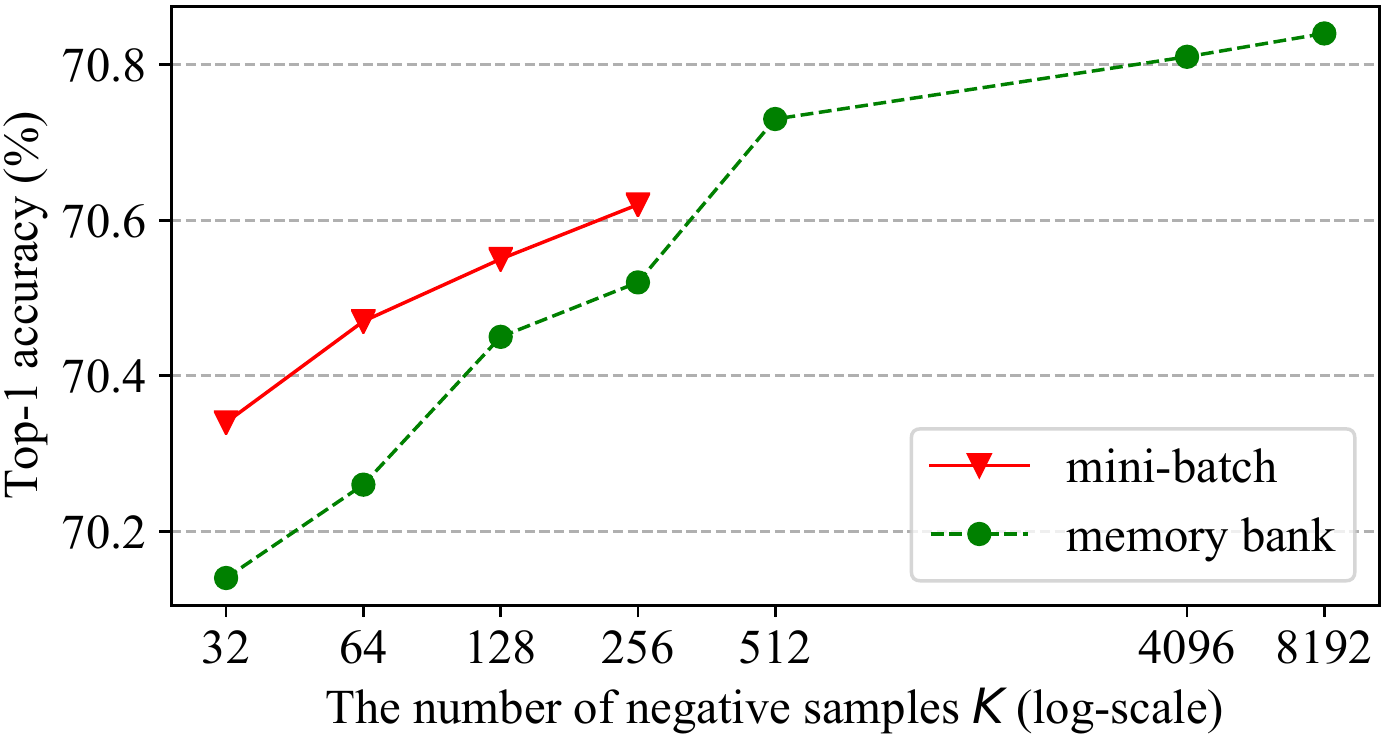}
		\caption{Top-1 accuracy (\%) over ResNet-18 on ImageNet.}
		\label{imagenet_neg}
	\end{subfigure}
	\caption{Comparison of two different methods of contrastive sample mining under various numbers of negative samples. The number of negatives is $K-2$ in the mini-batch-based mining (offset by two since the positive pair is retrieved from the same batch) and $K$ in the memory-bank-based mining.}
	\label{con_neg}
\end{figure} 

\textbf{Impact of different sample mining methods and the number of negative samples.} As shown in Fig.~\ref{con_neg}, we investigate two different methods of contrastive sample mining (referred to Section~\ref{Optimization}) with various numbers of negatives jointly. We conduct experiments on CIFAR-100 and ImageNet, representing the small- and large-scale datasets, respectively. First, both the two contrastive sample mining methods benefit from a larger $K$. And $K$ would saturate at a certain capacity for the memory bank. This observation is similar to some previous self-supervised works~\cite{wu2018unsupervised,he2020momentum} under the memory bank mechanism. It also suggests that more negative samples can guide the network to learn better feature representations. This is possibly
because the increased mutual information encourages more
knowledge transfer among networks, as claimed in Eq.~(\ref{mutual_infor}).

Generally, the mini-batch-based method outperforms the memory-bank-based counterpart under the same $K$ because those contrastive feature embeddings within the same mini-batch are \emph{consistent}~\cite{he2020momentum}. However, the batch size limits available negative samples in the mini-batch. As shown in Fig.~\ref{cifar_neg}, the performance upper-bound of the memory bank underperforms that of the mini-batch on the small-scale CIFAR-100, even if the batch size is small. In contrast, as shown in Fig.~\ref{imagenet_neg}, the memory-bank-based mechanism with more negatives breaks the performance upper-bound of the mini-batch-based counterpart on the large-scale ImageNet. We conjecture that this is because the required number of negative samples may be positively correlated with the dataset size for contrastive learning. ImageNet often needs more negative samples than CIFAR-100 to learn better features. Overall, we utilize mini-batch-based and memory-bank-based sampling methods for CIFAR-100 and ImageNet, respectively. And we use $K=126$ (\emph{i.e.} batch size$-2$) for CIFAR-100 and $K=8192$ for ImageNet.

\begin{table}[tbp]
	\centering
	\caption{Top-1 accuracy (\%) on ResNet-110 for CIFAR-100 and ResNet-18 for ImageNet under various contrastive embedding dimensions.}
	\begin{tabular}{c|cccc}  
		\hline 
		Dimension $d$&32&64&128&256 \\  
		\hline
		CIFAR-100&77.86$_{\pm 0.16}$&78.58$_{\pm 0.12}$&\textbf{79.05}$_{\pm 0.23}$&78.84$_{\pm 0.18}$ \\ 
		ImageNet&70.45&70.71&\textbf{71.13}&70.95 \\   
		\hline
	\end{tabular}
	\label{dimension}
\end{table}

\begin{figure}[tbp]  
	\centering
	\includegraphics[width=1\linewidth]{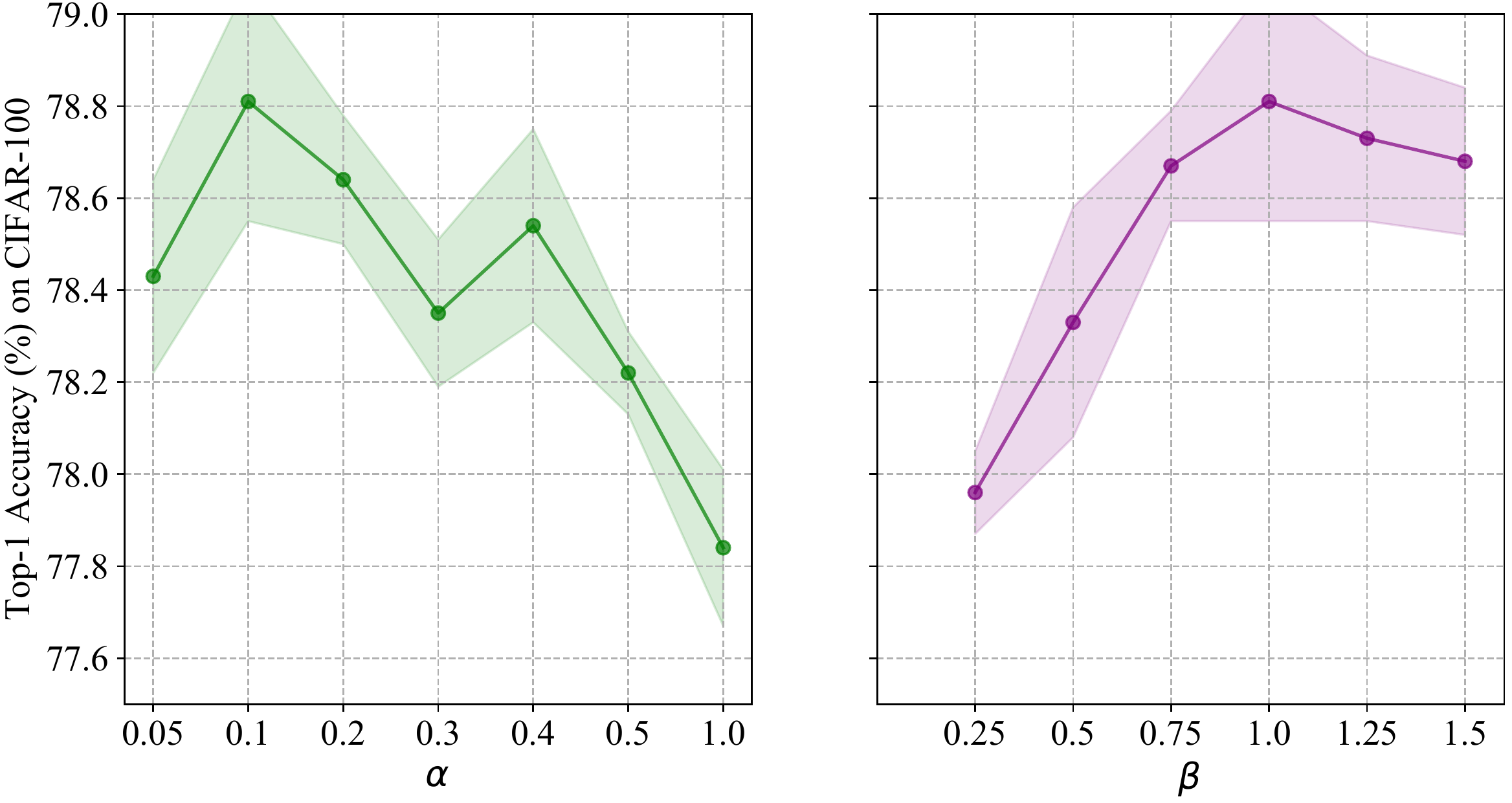}
	\caption{Sensitivity analyses of hyperparameters $\alpha$ and $\beta$.}
	\label{loss_weight}
\end{figure}

\begin{figure}[t]
	\centering 
	\begin{subfigure}[t]{0.48\textwidth}
		\centering
		\includegraphics[width=\textwidth]{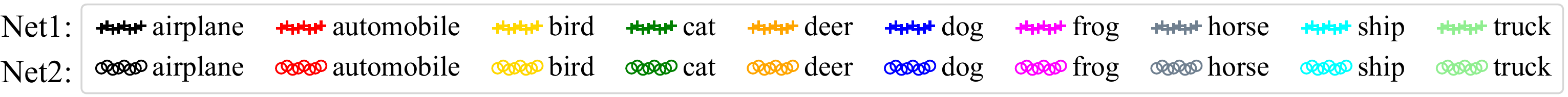}
		\caption{Legend of object classes from CIFAR-10.}
		\label{classes}
	\end{subfigure}
	\begin{subfigure}[t]{0.23\textwidth}
		\centering
		\includegraphics[width=\textwidth]{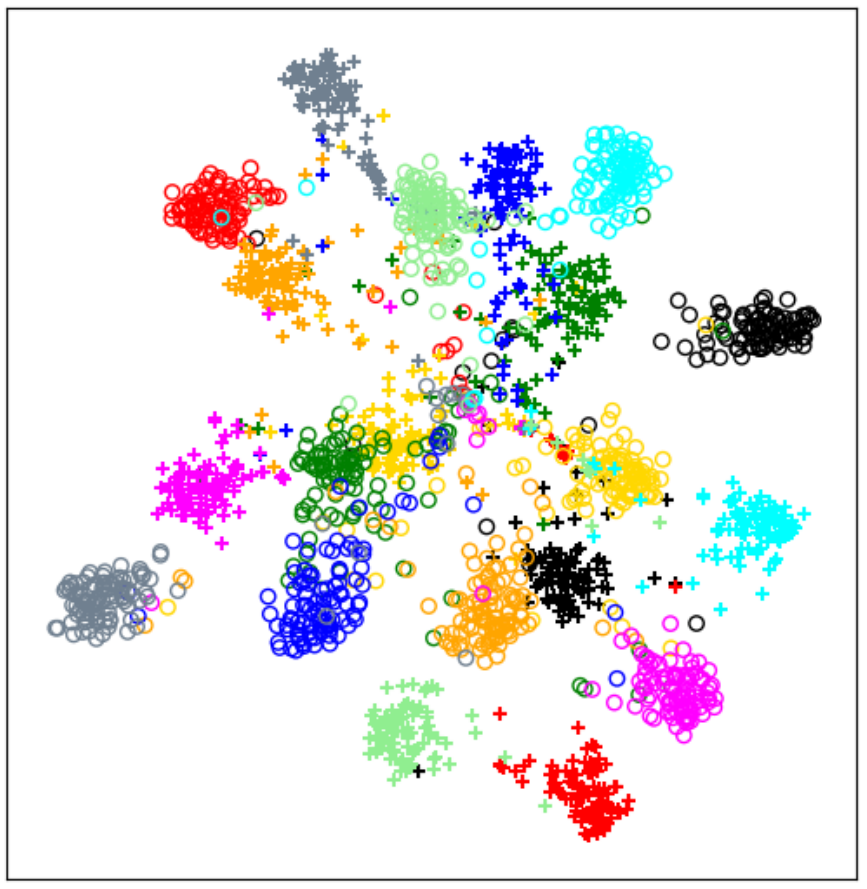}
		\caption{Independent training.}
		\label{Independent}
	\end{subfigure}
	\begin{subfigure}[t]{0.23\textwidth}
		\centering
		\includegraphics[width=\textwidth]{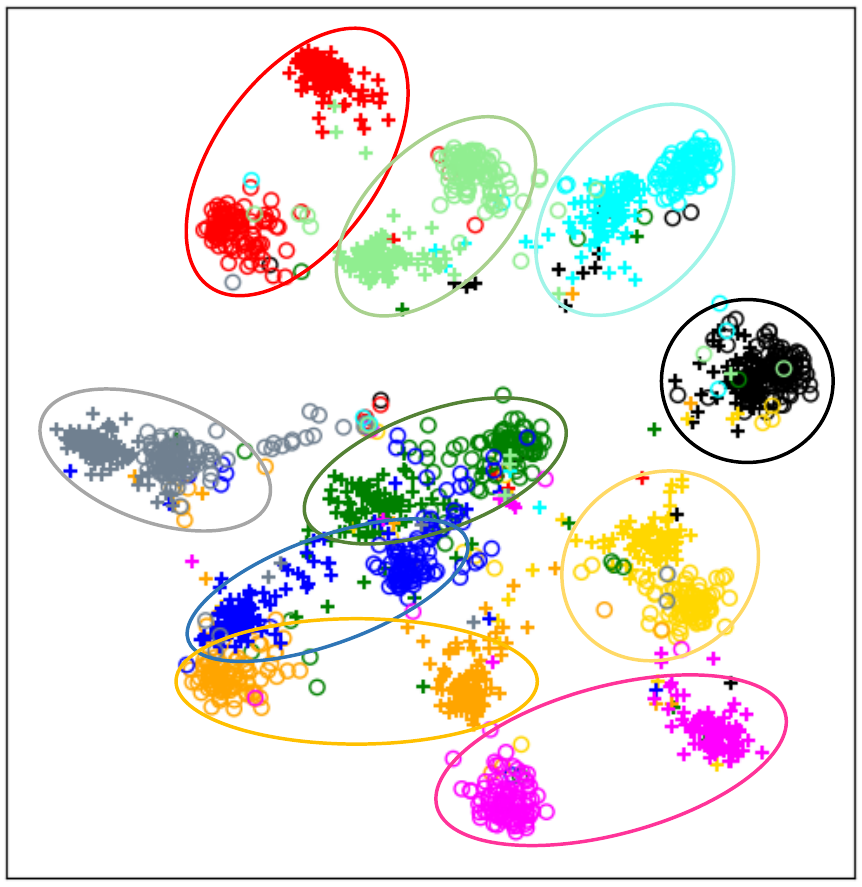}
		\caption{Our proposed MCL.}
		\label{mcl_tsne}
	\end{subfigure}
	\caption{T-SNE visualization of embedding spaces for two ResNet-32 (Net1 and Net2) with independent training (\emph{left}) and our MCL (\emph{right}) on CIFAR-10 dataset~\cite{krizhevsky2009learning} from~\cite{yang2022mutual}. The clusters in the same circle are from the same class.} 
	\label{tsne}
\end{figure} 
\textbf{Impact of contrastive embedding dimension.} We examine the impact of
embedding dimension $d$ to compute contrastive distributions for
L-MCL on CIFAR-100 and ImageNet. We start from $d=32$ to $d=256$ and find accuracy performance
steadily increases from 32, reaches plateaus at 128, and saturates at 256.

\textbf{Impact of loss coefficients $\alpha$ and $\beta$.} As shown in Fig.~\ref{loss_weight}, we investigate the impact of weight coefficients $\alpha$ and $\beta$, where $\alpha\in (0,1]$ and $\beta\in (0,1.5]$. We find that $\alpha\in [0.05,0.4]$ works well for
InfoNCE-based contrastive learning over supervised learning, but the performance decreases as $\alpha$ is greater than 0.4. We speculate that the larger percentage of contrastive learning may affect the original supervised task loss. Moreover, we also observe that KL divergence-based $\beta$ works well towards a weight of 1.0. This paper chooses $\alpha=0.1$ and $\beta=1.0$ as the best choices.

\textbf{Does MCL make networks more similar?} With the mutual mimicry by MCL, one may ask whether output embeddings of different networks in the cohort would get more similar. To answer this question, we visualize the learned embedding spaces of two ResNet-32 with independent training and MCL, as shown in Fig.~\ref{tsne}. We observe that two networks trained with MCL indeed show more similar feature distributions compared with the independent training baseline. However, the network trained by MCL achieves better accuracy performance than the baseline. This observation reveals that various networks using MCL can learn more common knowledge from others. Moreover, compared with the independent training, MCL can enable each network to learn a more discriminative embedding space, benefiting the downstream classification performance.

\section{Conclusion}
We propose Mutual Contrastive Learning, a method for online KD that trains a cohort of models from the perspective of contrastive representation learning. Layer-wise MCL is performed over the intermediate and final feature layers to facilitate sufficient knowledge interaction and maximize the mutual information between two networks. Experimental results across a broad range of visual recognition tasks, including image classification and object detection, demonstrate the superiority of layer-wise MCL to learn better features. This further leads to better performance than other popular logit-level online KD approaches. We hope our work can foster future research to take advantage of representation learning to improve online KD rather than only focusing on logit-level knowledge.

\ifCLASSOPTIONcompsoc
\section*{Acknowledgments}
\else
\section*{Acknowledgment}
\fi

This research work is supported by the National Key Research and Development Program of China under Grant No. 2021ZD0113602, the National Natural Science Foundation of China under Grant Nos. 62176014, the Fundamental Research Funds for the Central Universities.


%



\ifCLASSOPTIONcaptionsoff
  \newpage
\fi



%
\bibliographystyle{IEEEtran}
\bibliography{IEEEbib}

%

\begin{IEEEbiography}[{\includegraphics[width=1in,height=1.25in,clip,keepaspectratio]{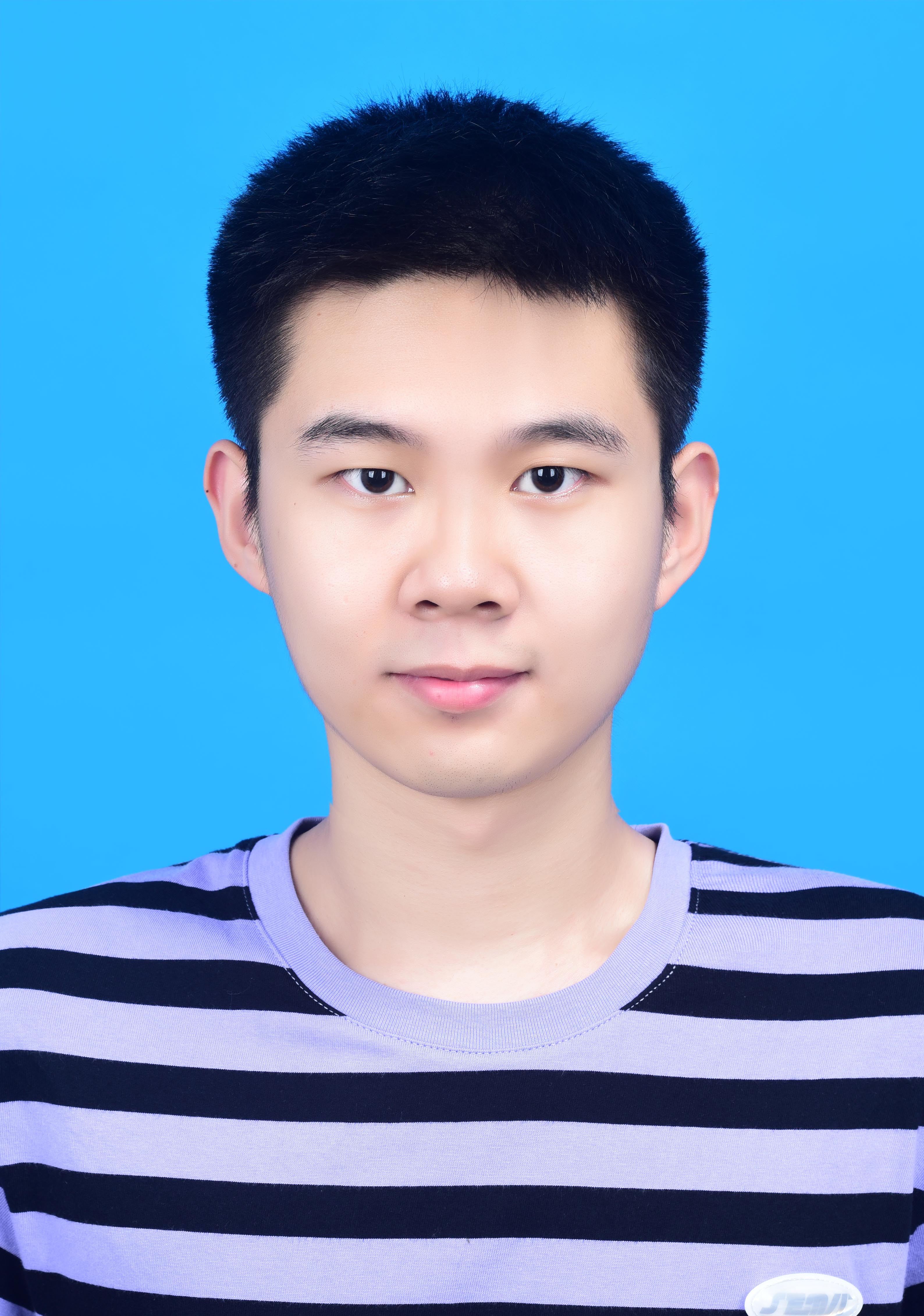}}]{Chuanguang Yang}
received the B.E. degree from Shandong Normal University, Jinan, China, in 2018. He is currently pursuing a Ph.D. degree with the Institute
of Computing Technology, Chinese Academy of
Sciences, China.  He has published papers on some prestigious refereed conferences such as CVPR, ICCV, ECCV, AAAI and IJCAI etc. His research interests include knowledge distillation, visual representation learning and image classification.
\end{IEEEbiography}

\begin{IEEEbiography}[{\includegraphics[width=1in,height=1.25in,clip,keepaspectratio]{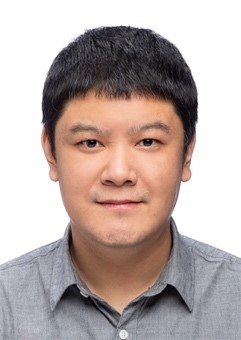}}]{Zhulin An}
	received the B.Eng. and M.Eng. degrees in computer science from Hefei University of Technology, Hefei, China, in 2003 and 2006, respectively and the Ph.D. degree from the Chinese Academy of Sciences, Beijing, China, in 2010.
	He is currently with the Institute of Computing Technology, Chinese Academy of Sciences, where he became a Senior Engineer in 2014. His current research interests include optimization of deep neural network and lifelong learning.
\end{IEEEbiography}

\begin{IEEEbiography}[{\includegraphics[width=1in,height=1.25in,clip,keepaspectratio]{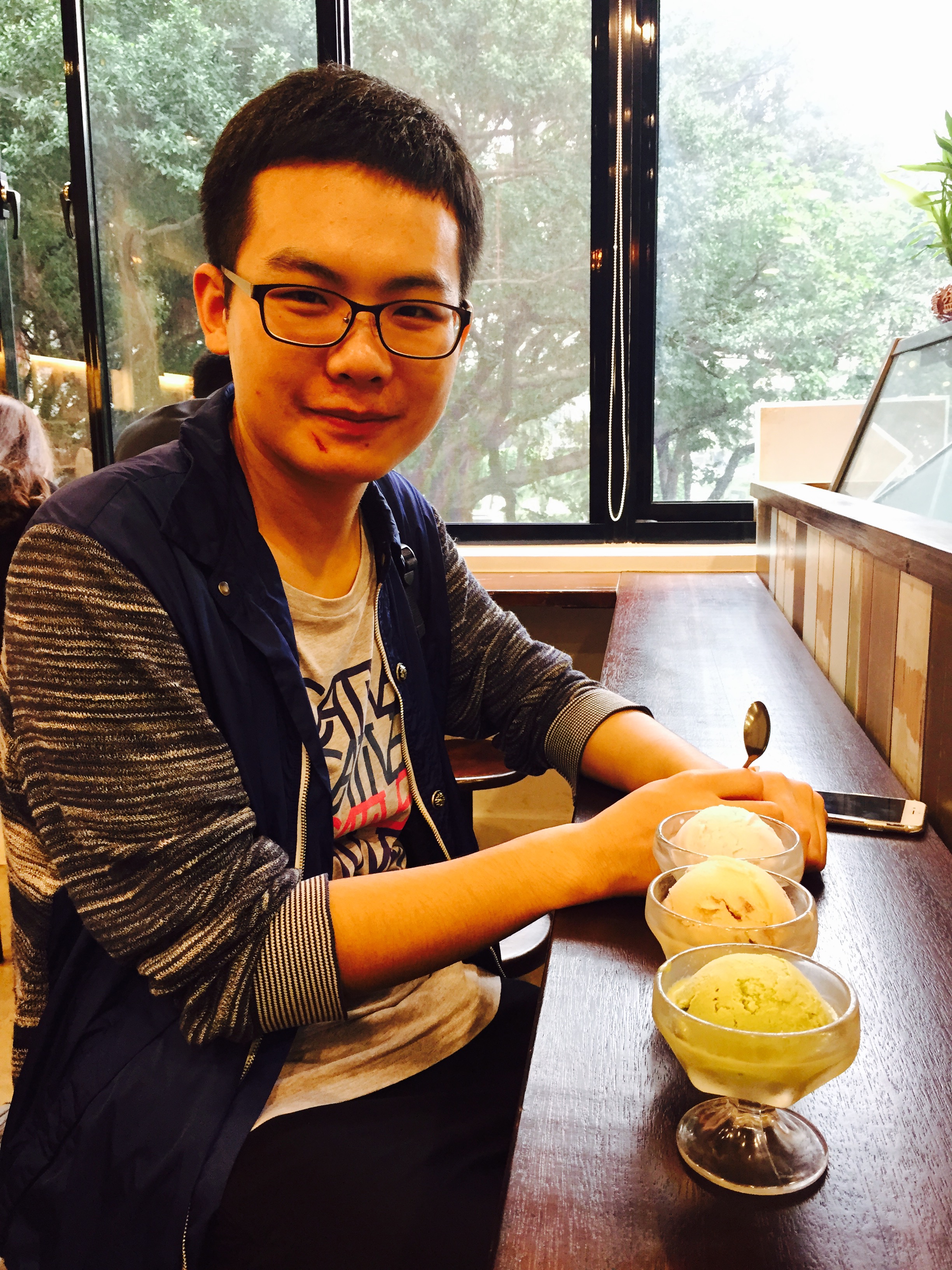}}]{Helong Zhou} received the B.E. degree from Wuhan University and the M.S. degree from National Taiwan University of Science and Technology. He is now a tech lead of Horizon Robotics in Beijing. His work focuses specifically on trajectory prediction, knowledge distillation, self-supervised learning and GAN.
\end{IEEEbiography}

\begin{IEEEbiography}[{\includegraphics[width=1in,height=1.25in,clip,keepaspectratio]{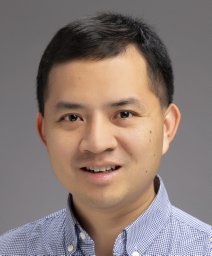}}]{Fuzhen Zhuang} is currently a Professor in Institute of Artificial Intelligence, Beihang University, Beijing, China. He received the Ph.D. degrees in computer science from the Institute of Computing Technology, Chinese Academy of Sciences, Beijing, China, in 2011. His research interests include machine learning, data mining, transfer learning, multi-task learning, recommendation systems and knowledge graph. He has published over 150 papers in the prestigious refereed journals and conference proceedings, such as Nature Communications, IEEE TKDE, Proc. of IEEE, TNNLS, TIST, KDD, WWW, SIGIR, NeurIPS, IJCAI, AAAI, and ICDE.
	
\end{IEEEbiography}

\begin{IEEEbiography}[{\includegraphics[width=1in,height=1.25in,clip,keepaspectratio]{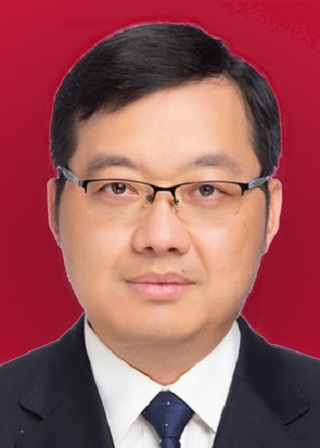}}]{Yongjun Xu}
	is a professor at Institute of Computing Technology, Chinese Academy of Sciences (ICT-CAS) in Beijing, China. He received his B.Eng. and Ph.D. degree in computer communication from Xi'an Institute of Posts \& Telecoms (China) in 2001 and Institute of Computing Technology, Chinese Academy of Sciences,
	Beijing, China in 2006, respectively. His current
	research interests include artificial intelligence
	systems, and big data processing.

\end{IEEEbiography}

\begin{IEEEbiography}[{\includegraphics[width=1in,height=1.25in,clip,keepaspectratio]{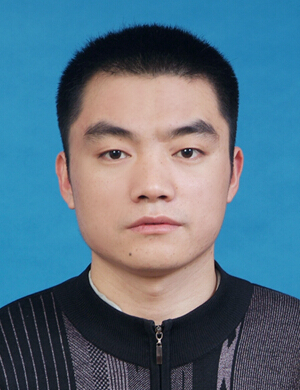}}]{Qian Zhang}
received the B.E. and M.S. degrees from Central South University, Changsha, China, in 2008 and 2011, respectively, and the Ph.D. degree in pattern recognition and intelligent systems from the Institute of Automation, Chinese Academy of Sciences, Beijing, China, in 2014. His current research interests include computer vision and machine learning.
	
\end{IEEEbiography}

%
%
%




\appendices
\section{Theoretical Insights of ICL}
\subsection{Proof of Maximizing the Lower bound of the Mutual Information}
Given the anchor embedding $\bm{v}_{a}^{0}$ from $f_{a}$ and contrastive embeddings $\{\bm{v}_{b}^{k}\}_{k=1}^{K+1}$ from $f_{b}$, we formulate the $(\bm{v}_{a}^{0}, \bm{v}_{b}^{1})$ as the positive pair and $\{(\bm{v}_{a}^{0}, \bm{v}_{b}^{k})\}_{k=2}^{K+1}$ as negative pairs. We consider the joint distribution $\mu(\bm{v}_{a},\bm{v}_{b})$ and the product of marginals $\mu(\bm{v}_{a})\mu(\bm{v}_{b})$. We define the distribution $q$ with an indicator variable $C$ to represent whether a pair $(\bm{v}_{a},\bm{v}_{b})$ is drawn from the joint distribution ($C=1$) or product of marginals ($C=0$):
\begin{align}
q(\bm{v}_{a},\bm{v}_{b}|C=1)=\mu(\bm{v}_{a},\bm{v}_{b}), \\ q(\bm{v}_{a},\bm{v}_{b}|C=0)=\mu(\bm{v}_{a})\mu(\bm{v}_{b}).
\end{align} 

Here, $C=1$ indicates the positive pair $(\bm{v}_{a}^{0}, \bm{v}_{b}^{1})$ while $C=0$ indicates a negative pair from $\{(\bm{v}_{a}^{0}, \bm{v}_{b}^{k})\}_{k=2}^{K+1}$, \emph{i.e.} $(\bm{v}_{a}^{0}, \bm{v}_{b}^{1})\sim \mu(\bm{v}_{a},\bm{v}_{b})$, $\{(\bm{v}_{a}^{0}, \bm{v}_{b}^{k})\}_{k=2}^{K+1}\sim \mu(\bm{v}_{a})\mu(\bm{v}_{b})$. For ICL, we often provide $1$ positive pair for every $K$ negative pairs. Therefore the prior probabilities of the latent variable $C$ are:
\begin{equation}
q(C=1)=\frac{1}{1+K},\  q(C=0)=\frac{K}{1+K}.
\end{equation}  
The class posterior of the pair $(\bm{v}_{a},\bm{v}_{b})$ belonging to the positive case ($C=1$) can be derived by Bayes’ rule:
\begin{align}
&q(C=1|\bm{v}_{a},\bm{v}_{b})\\
&=\frac{q(\bm{v}_{a},\bm{v}_{b}|C=1)q(C=1)}{q(\bm{v}_{a},\bm{v}_{b}|C=1)q(C=1)+q(\bm{v}_{a},\bm{v}_{b}|C=0)q(C=0)} \\
&=\frac{\mu(\bm{v}_{a},\bm{v}_{b})}{\mu(\bm{v}_{a},\bm{v}_{b})+K\mu(\bm{v}_{a})\mu(\bm{v}_{b})}
\end{align}
The log class posterior can be further expressed as follows:
\begin{align}
& \log q(C=1|\bm{v}_{a},\bm{v}_{b})\\
&=\log \frac{\mu(\bm{v}_{a},\bm{v}_{b})}{\mu(\bm{v}_{a},\bm{v}_{b})+K\mu(\bm{v}_{a})\mu(\bm{v}_{b})} \\
&=-\log(1+K\frac{\mu(\bm{v}_{a})\mu(\bm{v}_{b})}{\mu(\bm{v}_{a},\bm{v}_{b})}) \\
& \leq -\log(K)+\log\frac{\mu(\bm{v}_{a},\bm{v}_{b})}{\mu(\bm{v}_{a})\mu(\bm{v}_{b})}.
\end{align}

By computing expectations over the log class posterior, we can connect it to the mutual information as follows:
\begin{align}
& \mathbb{E}_{q(\bm{v}_{a},\bm{v}_{b}|C=1)}\log q(C=1|\bm{v}_{a},\bm{v}_{b})\\
& \leq -\log(K)+\mathbb{E}_{\mu(\bm{v}_{a},\bm{v}_{b})}\log\frac{\mu(\bm{v}_{a},\bm{v}_{b})}{\mu(\bm{v}_{a})\mu(\bm{v}_{b})}\\
& = -\log(K)+I(\bm{v}_{a},\bm{v}_{b})
.
\end{align}
In fact, we remark that the ICL loss $\mathcal{L}^{ICL}_{a\rightarrow  b}$ is the negative log class posterior of the positive pair:
\begin{equation}
\mathcal{L}^{ICL}_{a\rightarrow  b}=-\log q(C=1|\bm{v}_{a},\bm{v}_{b}).
\end{equation}
Therefore, we can connect $\mathcal{L}^{ICL}_{a\rightarrow  b}$ to the mutual information as follows:
\begin{align}
& \mathbb{E}_{q(\bm{v}_{a},\bm{v}_{b}|C=1)}\mathcal{L}^{ICL}_{a\rightarrow  b} \geq \log(K)-I(\bm{v}_{a},\bm{v}_{b}) \\
\Leftrightarrow\  &I(\bm{v}_{a},\bm{v}_{b})\geq \log(K)- \mathbb{E}_{q(\bm{v}_{a},\bm{v}_{b}|C=1)}\mathcal{L}^{ICL}_{a\rightarrow  b}.
\label{mutual_info}
\end{align}

\section{Pseudo-code of MCL}
\subsection{Algorithm of on CIFAR-100}
As shown in Algorithm~\ref{batch}, we perform batch-based mining in MCL for CIFAR-100-like small-scale datasets. 

\subsection{Algorithm of on ImageNet}
As shown in Algorithm~\ref{memory}, we perform memory-based mining in MCL for ImageNet-100-like large-scale datasets. 
\begin{algorithm}[tb]
	\caption{MCL on CIFAR-100 with batch-based mining}
	\begin{algorithmic}
		\label{batch}
		\STATE
		Initialize feature extractors $\{f_{m}\cup \zeta_{m}\}_{m=1}^{M}$
		\WHILE{$\{f_{m}\cup \zeta_{m}\}_{m=1}^{M}$ have not converged}
		\STATE 	Sample a mini-batch $\{\bm{x}^{(i)},{y}^{(i)}\}_{i=1}^{B}$, $B$ is batch size, $\bm{x}^{(i)}$ is the $i$-th sample and ${y}^{(i)}$ is the ground-truth label of $\bm{x}^{(i)}$.
		\STATE Given the $m$-th network, we infer feature embeddings  $\{\bm{v}^{(i)}_{m}\}_{i=1}^{B}$ from the input batch of $\{\bm{x}^{(i)}\}_{i=1}^{B}$.
		\STATE 	Given the $i$-th sample's embeddings $\bm{v}^{(i)}_{m}$ as the anchor, we retrieve positive embeddings $\mathcal{P}=\{\bm{v}_{m}^{(j)}\}_{j=1,j\neq i, y^{(j)}= y^{(i)}}^{B}$ and negative embeddings $\mathcal{N}=\{\bm{v}_{m}^{(j)}\}_{j=1,j\neq i, y^{(j)}\neq y^{(i)}}^{B}$. Due to the pre-defined class-aware sampler, $|\mathcal{P}|=1$ and $|\mathcal{N}|=B-2$ .
		\STATE Update $\{f_{m}\cup \zeta_{m}\}_{m=1}^{M}$  by optimizing loss $\mathcal{L}^{MCL}_{1\sim M}$
		\ENDWHILE
	\end{algorithmic}
\end{algorithm}

\begin{algorithm}[tb]
	\caption{MCL on ImageNet with memory-based mining}
	\begin{algorithmic}
		\label{memory}
		\STATE
		Initialize feature extractors $\{f_{m}\cup \zeta_{m}\}_{m=1}^{M}$
		\STATE Initialize $M$ memory banks $\{\mathbf{V}_{m}\}_{m=1}^{M}$ with random unit vectors
		\WHILE{$\{f_{m}\cup \zeta_{m}\}_{m=1}^{M}$ have not converged}
		\STATE 	Sample a mini-batch $\{\bm{x}^{(i)},{y}^{(i)}\}_{i=1}^{B}$, $B$ is batch size, $\bm{x}^{(i)}$ is the $i$-th sample and ${y}^{(i)}$ is the ground-truth label of $\bm{x}^{(i)}$.
		\STATE Given the $m$-th network, we infer feature embeddings  $\{\bm{v}^{(i)}_{m}\}_{i=1}^{B}$ from the input batch of $\{\bm{x}^{(i)}\}_{i=1}^{B}$.
		\STATE 	Given the $i$-th sample's embeddings $\bm{v}^{(i)}_{m}$ as the anchor, we randomly retrieve $1$ positive embedding $\bm{v}_{m}^{1}$ and $K$ negative embeddings $\{\bm{v}_{m}^{k}\}_{k=2}^{K+1}$ from the $m$-th memory queue $\mathbf{V}_{m}$. 
		\STATE Update $\{f_{m}\cup \zeta_{m}\}_{m=1}^{M}$  by optimizing loss $\mathcal{L}^{MCL}_{1\sim M}$
		\STATE Enqueue the mini-batch embeddings $\{\bm{v}^{(i)}_{m}\}_{i=1}^{B}$ to $\mathbf{V}_{m}$
		\STATE Dequeue early embeddings from $\mathbf{V}_{m}$
		\ENDWHILE
	\end{algorithmic}
\end{algorithm}

\section{Overview of The Proposed Framework}
\begin{figure*}[htbp]  
	\centering  	\includegraphics[width=1\linewidth]{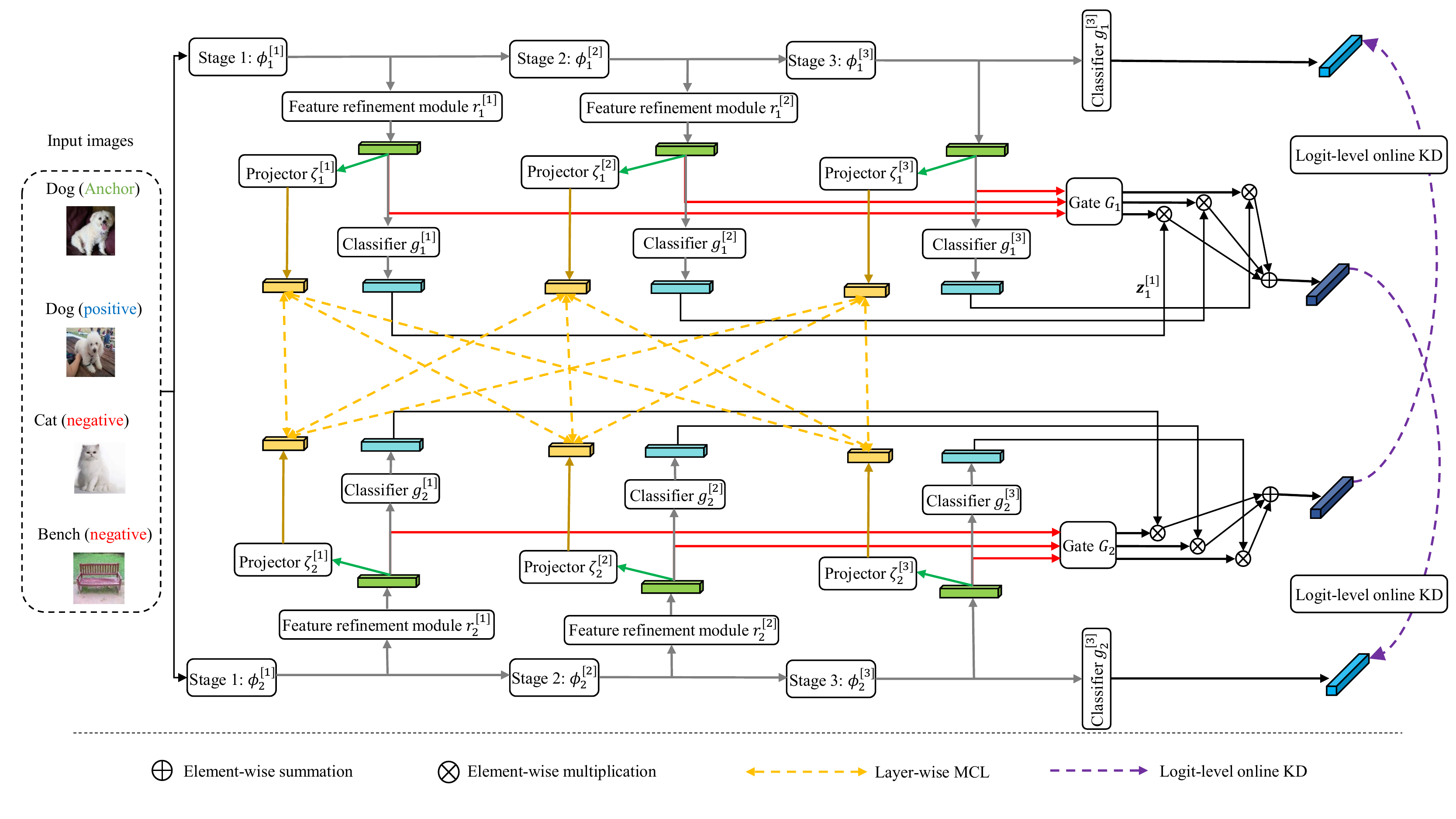}
	\caption{Overview of our proposed layer-wise MCL and logit-level online KD methods over two networks $f_1$ and $f_2$.} 
	\label{overview}
\end{figure*}

As shown in Fig.~\ref{overview}, we illustrate the overview of the proposed framework, including feature-based layer-wise MCL and logit-based distillation.

\end{document}